\documentclass[12pt,a4paper]{article}
\usepackage[left=2cm,right=2cm,top=2cm,bottom=2cm]{geometry}
\usepackage[UKenglish]{isodate}
\usepackage[pagewise]{lineno}

\usepackage{setspace}
\usepackage[utf8]{inputenc}
\usepackage{float}
\usepackage{hyperref}
\usepackage{amsmath}
\usepackage{mathtools}
\usepackage{algorithm}
\usepackage{algorithmic}
\usepackage{graphicx}
\usepackage{booktabs}
\usepackage{braket}
\usepackage{amsfonts}
\usepackage[english]{babel}
\usepackage{multirow}
\usepackage{lscape} 
\usepackage{afterpage}
\usepackage{pdflscape}
\usepackage{everypage}
\usepackage{amsthm}
\usepackage[table]{xcolor}
\usepackage{capt-of}
\usepackage{floatpag}
\usepackage{verbatim}
\usepackage{ifoddpage}

\usepackage{caption}
\newtheorem{theorem}{Theorem}[section]
\newtheorem{lemma}[theorem]{Lemma}
\newtheorem{proposition}[theorem]{Proposition}
\newtheorem{corollary}[theorem]{Corollary}
\renewcommand{\baselinestretch}{1.45}

\newcommand{\Lpagenumber}{\ifdim\textwidth=\linewidth\else\bgroup
  \dimendef\margin=0 
  \ifodd\value{page}\margin=\oddsidemargin
  \else\margin=\evensidemargin
  \fi
  \raisebox{\dimexpr -\topmargin-\headheight-\headsep-0.5\linewidth}[0pt][0pt]{%
    \rlap{\hspace{\dimexpr \margin+\textheight+\footskip}%
    \llap{\rotatebox{90}{\thepage}}}}%
\egroup\fi}
\AddEverypageHook{\Lpagenumber}%

\usepackage{tcolorbox}
 
\usepackage{natbib}
\begin{document}

\title{\vspace*{-0cm}From interpretability to inference:\\an estimation framework for universal approximators}
\author{Andreas Joseph}
\date{\vspace*{-.3cm}Bank of England\thanks{{\it Disclaimer:} The views expressed in this work are not necessarily those of the Bank of England or one of its committees. All errors are mine. {\it Contact:} Bank of England, Advanced Analytics Division (Threadneedle Street, London EC2R 8AH, UK). Email: \href{mailto:andreas.joseph@bankofengland.co.uk}{andreas.joseph@bankofengland.co.uk}. {\it Acknowledgement:} 
Many thanks to the multitude of people who supported and commented on this work over several years; in particular to Tunrayo Adeleke-Larodo, David Bholat, Philippe Bracke, David Bradnum, Marcus Buckmann, Mingli Chen, Victor Chernozhukov, Sinem Hacioglu, George Kapetanios, Anton Korinek, Michele Lenza, Evan Munro, Milan Nedeljkovic, Eric Renault, Paul Robinson, Arthur Turrell, Hal Varian, Eryk Walczak, and the participants of numerous conferences, workshops, and seminars. Special thanks to the Behavioural Insights Team for making the data for the empirical case study available. 
{\it Note:} 
The current paper is a revised and extended version of papers previously titled ``Shapley regressions: A framework for statistical inference on machine learning models'' and ``Parametric inference with universal function approximators''. The Online Appendix can be found \href{https://www.dropbox.com/scl/fi/rj83iemdixg80heu9nr5n/Shapley_inference_v8_AE_ONLINE.pdf?rlkey=3230kx5ilcqf4icxcpbq5e55i&dl=0}{here}.
}\\[.3cm]
	\cleanlookdateon \today
%
		}
%

\maketitle

\vspace*{-1cm}
\begin{abstract}
We present a novel framework for estimation and inference with the broad class of universal approximators. Estimation is based on the decomposition of model predictions into Shapley values. Inference relies on analyzing the bias and variance properties of individual Shapley components. We show that Shapley value estimation is asymptotically unbiased, and we introduce Shapley regressions as a tool to uncover the true data generating process from noisy data alone. The well-known case of the linear regression is the special case in our framework if the model is linear in parameters. We present theoretical, numerical, and empirical results for the estimation of heterogeneous treatment effects as our guiding example.\\[.2cm]
{\bf Keywords:} statistical learning,  Shapley values, statistical inference, treatment effect estimation.\\[.2cm]
{\bf JEL codes:} C14, C31, C45, C52, C71, E52.\\[.3cm]
\end{abstract}

\clearpage

\section{Introduction}
\label{sec:intro}

This paper connects the literature on interpretable machine learning with that of estimation and inference in statistics and econometrics. Models from machine or statistical learning, like artificial neural networks (ANN), or tree-based models like random forests (RF), are increasingly being used to address a wide range of problems in economics and finance.\footnote{Examples include demand estimation \citep{Bajari2015demand,Guha2019sales}, the modeling of financial distress \citep{Kim2010credit,Schalck2021sme} or risk \citep{bracke2019risk,Mashrur2020risk,Chronopoulos2023quantile}, asset pricing \citep{gu2020assets,Bianchi2020premia,Kapetanios2023assets}, macroeconomic forecasting \citep{Nakamura2005inflation,Coulombe2022forecasting,joseph2024cpi}, financial crisis prediction \citep{Ward2017crisis,bluwstein2023crisis}, the estimation of treatment effects \citep{athey2016forest,chernozhukov2016double,wager2018treatment,Chernozhukov2022automatic}, or the solution of structural models \citep{MALIAR2021deep,Norets2021neural,kase2022estimate,villaverde2023Inequality}. We use the terms statistical and machine learning interchangeable here.} With the notable exception of treatment effect estimation, little is generally known about the estimation and inference properties of statistical learning models. For instance, the output obtained from different machine learning models can substantially differ across models and between tasks \citep{Fernandez2014} without a clear understanding of the differences between models post optimization. We address this situation by providing a generic framework for estimation and inference for the broad class of universal approximators. This class of models also lies at the heart of modern machine learning advances based on deep learning \citep{goodfellow2016deep}. We provide a comprehensive theory and the practical tools to statistically analyze model outputs.\\
The framework is based on the analysis of the decomposition of model predictions into Shapley values \citep{Strumbelj2010shapley}, a key innovation in interpretable machine learning. This decomposition makes an analogy between the prediction of a model and the payoff from a cooperative game. In the latter, an established solution to the problem of attributing a share of the payoff to a player is its Shapley value \citep{shapley1953shap,young1985shap}. For a model, this is the predictive share attributed to an input. The Shapley value decomposition of model predictions has become a leading approach to explain machine learning models, because it inherits the mathematical properties from its game theoretic origin, and because several popular model explanation approaches have been shown to map into a Shapley value decomposition \citep{lundberg2018shap}. However, this decomposition only is descriptive in the sense that there exists no statistical theory around it. For example, it would be of interest if and how the Shapley decomposition can be connected to the underlying data generating process (DGP). The current paper addresses this gap between model interpretability and statistics.\\
We show that estimation with Shapley values is asymptotically unbiased in fairly broad empirical settings, and can, as such, be used to uncover the true DGP. The presented theory has several appealing properties. First, estimation and inference with Shapley values reduces to the well known case of the linear model, i.e.\ the analysis of regression coefficients, if the model is linear in parameters. And second, the asymptotics can be used to analyze the statistical learning process itself. For instance, we suggest simple tests to assess whether a signal from a model's Shapley decomposition is likely to contribute to the DGP of interest, or whether it is noise; and on how close that signal is to the unknown true value (bias assessment).\\
In a nutshell, the proposed framework consists of three parts. One, the estimation of effects using the components of a generalized Shapley-Taylor decomposition of model predictions \citep{mukund2019shapley_taylor}. Two, inference through the quantification of sampling and sample split uncertainty using bootstrap approaches adapted to training-test situations common in statistical learning. And third, the assessment of the information content and bias of individual Shapley components using an auxiliary linear regression.\\
While the proposed framework can be applied to a large set of problems, the estimation of heterogeneous treatment effects serves as our guiding theoretical and empirical example. This has been one of the main applications of statistical learning models in economics and econometrics (see \cite{athey2016forest,chernozhukov2016double,wager2018treatment} for seminal contributions). The Shapley value estimation of treatment effects can be achieved via a {\it direct} approach where only the response surface is modeled. That is, there is no need for orthogonalization and the estimation of propensity scores, which we call the {\it indirect} approach. However, we do not see the this direct estimation approach as a substitute, but rather the tools provided here can serve as a supplements to indirect estimation. For instance, using either approach Shapley values can be used to derive a general treatment function. This expresses treatment effects as a potentially nonlinear function of covariates of interest. This not only allows one to measure heterogeneity of treatment effects, but also to potentially identify treatment channels. This subsequently may help to increase the external validity of treatment interventions \citep{deaton2018rct}.\\
We consider a numerical case study and revisit a real-world experiment. The numerical case study demonstrates how the proposed theory can be used to recover the true (heterogeneous) treatment effects asymptotically while comparing different off-the-shelf statistical learning models. All models recover the true DGP perfectly from noisily observed raw inputs given enough data. At the same time, differences in learning outcomes are shown to align with the different properties of the used models.\\
The empirical case study investigates the effects of an information treatment on the effectiveness of central bank communication \citep{BHOLAT2019}. In a randomized control trial, participants are either shown a text (control) or an graphical (treatment) summary of an actual monetary policy announcement of the Bank of England. The understanding of either material is assessed by a comprehension test. We estimate and compare (heterogeneous) treatment effects using a standard linear model (no heterogeneity), established approaches from statistical learning \citep{wager2018treatment,kunzel2019meta}, and direct estimation with off-the-shelf models. We decompose the predicted treatment effects from either estimation approach into its Shapley components and compare the results using the tools presented in this paper.\\
We find that most machine learning models learn the average treatment effects well, i.e.\ that Shapley value estimates are in line with the linear unbiased model. However, the distributions of the estimated treatment heterogeneity can vary substantially, between models and estimation approaches. We use inference on components of the Shapley treatment function to compare the learned signals. We find that some models, especially those based on support vector machines (SVM) learn a nonlinear relationship between treatment outcomes and age. The information treatment seems to be the more effective the older participants are. Because the information treatment was designed to relate to lived experiences, this could suggest that this kind of intervention is the more effective the more `life experience' a person has, and that different interventions may be needed to better reach younger audiences, for which we observe considerably smaller treatment effects. Furthermore, the positive treatment effect levels off at between 60 and 70 years of age. This is an example of a nonlinear relation which our approach helps to uncover. A more nuanced picture could then be that life experience matters up to a point within this setting.\\
%
The remainder of the paper is structured as follows. Section\ \ref{sec:literature} reviews the related literature on interpretable machine learning, and estimation and inference with statistical learning models. Section\ \ref{sec:method} introduces the notation, the assumptions, the training and test setting, model Shapley values, and the Shapley-Taylor decomposition of model predictions. Section\ \ref{sec:main} introduces the main theoretical results for estimation and inference with Shapley values. Section\ \ref{sec:applications} presents the numerical and empirical case studies on the estimation of heterogeneous treatment effects. Section\ \ref{sec:conclusion} concludes with a discussion. Proofs are given in the \hyperref[sec:app]{Appendix}.

\section{Related literature}
\label{sec:literature}

This paper brings together the two largely separate literatures of interpretable machine learning, and estimation and inference with machine learning models. The former is primarily concerned with addressing the black box critique of these models, and mostly originated in computer science. The latter is concerned with non- or semi-parametric estimation, and was mostly developed in econometrics.\\
Regarding interpretability, a primary concern with the use of machine learning models is the black box critique. That is, given an optimized, or trained, model, its input-output relations are not directly accessible even if one has full access to the model. This is because the internal parameter space of a model generally is degenerate removing any intrinsic meaning of individual parameters (see next section for details).
The attribution of importance scores to the individual predictors entering a model is a general approach to address this problem (see \cite{molnar2019xai} for an overview). These scores can be global or local. Global scores assign a single value to each variable across the input domain. A prominent example is variable importance for tree-based models \citep{friedman2001elements,Kazemitabar2017importance} Local measures provide scores for individual predictions (e.g.\ LIME \citep{Ribeiro2016lime} or Shapley values \citep{Strumbelj2010shapley}). Local measures can always be aggregated to give a global measure, hence, they carry more information content. \cite{lundberg2018shap} show that Shapley values unify several local explanation measures. Together with the appealing analytical properties stemming from their game theoretic origin, it can be argued that Shapley values are the preferred measure to assess the importance of variables in a wide range of supervised learning settings. However, \cite{sundararajan2019shapley} showed that, despite their axiomatic definition, the operationalization of Shapley values often can render uniqueness results meaningless, and can lead to counter-intuitive attributions. We provide a condition, which usually is fulfilled in empirical settings, to alleviate their concerns.\footnote{This supplementary result is stated in the Online Appendix where we discuss the properties and computation of model Shapley values which are not the focus of our main study.}\\
Regarding estimation and inference, machine learning models have mostly been used to estimate treatment effects in different settings. Major applications are the estimation of effects in the presence of high-dimensional nuisance parameters \citep{chernozhukov2016double,Chernozhukov2022automatic}, or the estimation of heterogeneous effects using modified tree models \citep{athey2016forest,wager2018treatment}. A key insight from this literature is that we need to account for biases stemming from regularization and overfitting of statistical learning models. The former is addressed by the construction of orthogonalized scores, and the latter by using cross-fitting which we will rely on as well. The orthogonalized scores are constructed from separate supervised prediction problems for which statistical learning model are well suited for, such as the response or treatment probability in an high-dimensional setting. Furthermore, it has been shown that under relatively permissive conditions valid confidence intervals can be computed.\\
In this vein, meta-learners \citep{kunzel2019meta} combine potentially different machine learning models to model different response surfaces, like the outcomes of the treated and control groups based on the controls, or the propensity score. These separate prediction models are then combined, depending on the type of learner, to estimate heterogeneous treatment effects. A potential advantage of meta-learners is that they allow for a flexible combination of different models for each component entering the estimation process. However, this flexibility comes at the cost of inference, as they typically do not offer valid confidence intervals. We will show how Shapley value estimation and inference can address this gap owing to its generality.\\
Yet another approach develops estimation and inference properties for specific models. \cite{Farrell2021deep} establish  nonasymptotic bounds for commonly used ANN types and nonparametric regression problems, like least squares or logistic regressions. This work is extended to a panel setting by \cite{chronopoulos2023deep}. All of the above approaches assume that learning is possible, i.e.\ that the nonparametric approximation of quantities of interest is possible. However, there are situations where statistical learning fails due to impossibilities in nonparametric convergence \citep{stone1982}. Such a situation is addressed in \cite{chernozhukov2017generic}, where the authors provide tools for the estimation of aspects of treatment effects despite impossibilities in learning due to dimensional restrictions. While we will assume learnability, this work points to interesting extensions of the current work.

\section{Methodological background}
\label{sec:method}

\subsection{Modelling setting and notation}
\label{sec:notation}

We consider the common case of modeling a noisy signal or target, $y_i\,=\,f(x_i;\alpha)\,+\,\eta_i$, with $f: D_0\subset \mathbb{R}^{n_0}\mapsto T\subset\mathbb{R}^r$ being the data generating process (DGP) of interest and $\eta\in\mathbb{R}^r$ an irreducible noise component with zero mean. The DGP is assumed to be continuous within $T$ and $f$, $x$, $\eta$, and as such $y$, are assumed to have finite variance. The vector $\alpha\in\mathbb{R}^{p}$ describes the parameterization of the DGP. We observe the data $x\subset\Omega\subset\mathbb{R}^{m\times n}$ with $n$ being the number of features or variables and $m$ the number of i.i.d.\ observations. We assume no omitted variables, i.e.\ $x$ contains all variables present in the DGP $f$ while we may unknowingly observe noise variables unrelated to the DGP, such that $n\geq n_0$, where $n_0$ is the number of variables entering $f$.\\
The nonparametric model $\hat{y}=\hat{f}(x;\theta): D\subset \mathbb{R}^{n}\mapsto T\subset \mathbb{R}^r$, where $\theta\in\mathbb{R}^{q}$, and $q\rightarrow\infty$ as $m\rightarrow\infty$ is allowed, and $var(\hat{f})<\infty$. This gives the generic modeling setting, 
\begin{equation}\label{eq:modelling}
y_i\,=\,\hat{f}\big(x_i;\theta)\,+\,\epsilon_i\,,\nonumber
\end{equation}
with the mean-zero residual vector $\epsilon_i\in\mathbb{R}^{r}$. We only consider the one-dimensional case $r=1$ without loss of generality. The optimization problem for our models is to minimize an empirical risk $R^e=\frac{1}{m}\sum_{i=1}^m||\hat{y_i}-y_i||=\frac{1}{m}\sum_{i=1}^m||\epsilon_i||$, where $||\cdot||$ is a distance measure depending on the model. This describes a regression setting, while most aspects discussed here can be straightforwardly transferred to classification problems by varying the risk function \citep{vapnik1999learning}.\\ 
The main difference between the sets of parameters $\alpha$ and $\theta$ is that the former identify the DGP, while the latter may be degenerate in the sense that different configurations of $\theta$ can describe the same model.\\
We use the index convention that $i,j\in \{1,\dots,m\}$ refer to individual observations (rows of $x$) and $k,l\in \{1,\dots,n\}$ to variable dimensions (columns of $x$). No index refers to the whole dataset. Sample averages are barred. Estimated quantities are hatted, except decompositions or their components $\phi/\Phi$ for simplicity. Primed inputs, e.g.\ $x'$, refer to variable sets out of the set of possible variable coalitions $\mathcal{C}(x)$, where each variable is allowed to enter at most once. We mostly refrain from using set braces for a simplified notation, and $|x'|\leq n$ is the number of variables in $x'$ (also indexed by $k,l$ if needed). Super-scripts $S$ refer to `Shapley-related' quantities, which will be clear from the context. Star-quantities ($\phi^{\star}/\Phi^{\star}$) refer to (unobserved) true values.

\subsection{The key assumption of learning}
\label{sec:learning}

The main assumption which many of our results are based on is that statistical learning is possible. That is, the empirical risk $R^e$ converges uniformly in probability to the minimally achievable loss given by the irreducible error $\eta$ with a rate $\xi_{\mathrm{ml}}$, i.e.\ $R^e\sim m^{-\xi_{\mathrm{ml}}}$. Concretely, the expected value of model predictions convergences uniformly in probability to the true DGP
\begin{equation}\label{eq:err_consist}
\lim_{m_{\mathrm{train}} \to \infty} \mathbb{E}_{\mathrm{train}}\big[||f(x_i)-\hat{f(}x_i)||\big]=0\;,\;\forall x_i \in x_{\mathrm{test}}\,.
\end{equation}

\noindent The expectation $\mathbb{E}_{\mathrm{train}}$ refers to any independent training data $x_{\mathrm{train}}\subset\Omega$ of size $m_{\mathrm{train}}$ (training sample), while model evaluation is based on an separate hold-out sample $x_{\mathrm{test}}\subset\Omega$ (test sample). Property (\ref{eq:err_consist}) is called {\it error consistency}, because the expected error $\epsilon$ vanishes everywhere despite the presence of the irreducible noise $\eta$. The model $\hat{f}$ is a {\it universal approximator} if (\ref{eq:err_consist}) holds.\footnote{Many popular machine learning models, like different types of ANN, RF, and SVM have been shown to be universal approximators. See for example \cite{cybenko1989universal,geman1992consistent,Farago1993nn,Steinwart2002svm,
steinwart2007,christmann2008svm,biau2012rf,biau2014rf,andoni2016poly} and references therein.}
This will be our main prerequisite:\\[.2cm]
{\it Assumption 1:} Model $\hat{f}$ is a universal approximator in the sense of (\ref{eq:err_consist}).\\[.2cm]
We need the train-test split of the data, such that $\mathbb{E}_{\mathrm{train}}[\epsilon_i]=0,\;\forall x_i \in x_{\mathrm{test}}$. This can be seen as follows. Assume that a training observation $i$ influences its own target prediction $\hat{f}(x_i)$ by some value $|\delta_i| > 0$ through the optimization process. That is, including $i$ in the training data moves its predicted value by $\delta_i$. Then, we have an unknown relationship $\epsilon_i(\delta_i)$ with $\mathbb{E}_{\mathrm{train}}[\epsilon_i(\delta_i)]\neq 0$ in general, i.e.\ there is a wedge created by in-sample evaluation. Thus, all model expectations will be taken as in (\ref{eq:err_consist}), and we will drop the $\mathrm{train}$- and $\mathrm{test}$-subscripts in most instances unless explicitly needed for clarity.

\subsection{Model training and testing}
\label{sec:train_test_setting}

Many of our results will depend on appropriate splits of the data into training and test data sets. A sample-efficient way for this is $K$-fold cross-fitting \citep{friedman2001elements}: assuming for simplicity that $m$ is divisible by $K\geq 2$, we divide the data into $K$ equally sized partitions. These are used as $K$ test data sets, with the remaining $K-1$ partitions in each case being used for training. Iterating through these training and testing partitions allows to obtain valid test scores for all observations. While being sample efficient, this approach introduces additional sample split uncertainty which we need to account for. To be able to do so, we evaluate results over $R$ random cross-fitting realization. That is, we obtain $R$ estimates at each point $x_i$ for every quantity of interest and take the measured variation into account.\\
An important question is what $K$ should be. Some of our main results will depend on $\xi_{\mathrm{ml}}\geq \frac{1}{2}$. However, we often have $\xi_{\mathrm{ml}}<\frac{1}{2}$ in nonparametric learning settings (\cite{stone1982}). The solution is to choose $K$ large enough to increase the speed of nonparametric learning using the training partitions relative to the parametric rate of $\xi_{p}=\frac{1}{2}$ on the test partitions. In particular, we can set 
%
\begin{equation}\label{eq:k}
K \,\geq\, \Big\lceil m^{1-2\xi_{\mathrm{ml}}}\Big\rceil+1\,\equiv\,\overline{K}\,, 
\end{equation}
where equality leads to the parametric rate for quantities evaluated on the test sets if $m^{1-2\xi_{\mathrm{ml}}}$ is an integer. To provide some intuition for the above expression, we can set $\xi_{\mathrm{ml}}=\xi_{p}=\frac{1}{2}$. This leads to $\overline{K}=2$, which is the smallest possible number of folds, splitting the data into one training and one test partition. Now, defining $\underline{K} \equiv \lfloor m^{1-2\xi_{\mathrm{ml}}}\rfloor+1$, we arrive at
\begin{lemma} (super-convergence) 
\label{math:superconv} Let $\xi_{\mathrm{ml}}\leq\frac{1}{2}$ be the convergence rate for training $\hat{f}$. If $K>\underline{K}$, the variance of an estimator $E(\hat{f},x_i)$ linear in $\hat{f}$ coming from sample splitting vanishes asymptotically relative to that of a classical $\sqrt{m}$-estimator, e.g.\ that of a linear regression coefficient. The \hyperref[app:proof_superconv]{proof} is given in the Appendix.
\end{lemma}
%
\noindent  The intuition behind Lemma\ \ref{math:superconv} is that, given a large enough $K$, the difference between models trained on different training splits vanishes relative to sample variance which is fixed at the parametric rate. This can have practical implications. Given a large data set, a large $K$ reduces the effects of sample split uncertainty, making the choice of a single cross-fitting realization appropriate ($R=1$), thereby reducing computational needs.\\[.2cm]
{\it Assumption\ 2:} The training-test setting is always such that the model $\hat{f}$ converges at least with the parametric rate, i.e.\ $K \geq \overline{K}\;\Rightarrow\;\xi_{\mathrm{ml}} \geq \xi_p=\frac{1}{2}$ in (\ref{eq:err_consist}).\\[.2cm]
Most models need hyperparameter tuning, such as setting regularization parameters or the ANN network size. We will use $K'$-fold cross-validation, which follows the same principle as $K$-fold cross-fitting on a separate validation data set  (simulations) or nested cross-validation on the training data set in empirical case studies. Standard values of $K'$ are five or ten.\footnote{The uncertainty stemming from cross-validation is empirically captured by measures of the sample split uncertainty.}

\subsection{Model Shapley values}
\label{sec:shapley_values}

The linear model $\hat{f}(x_i)=x_i\theta=\sum_{k=0}^{n}x_{ik}\theta_{k}$, with $\theta_{0}$ the intercept and $x_{i0}=1$, is special in the sense that it provides local and global estimation and inference at the same time.
The coefficients $\theta$ describe {\it local} effects via the sum of the product of input components and coefficients at each point $x_i$. At the same time, the coefficient vector $\theta$ determines the orientation of the {\it global} model hyperplane with constant slope in each direction of the input-output space.\\
The linear model belongs to the class of local additive variable attributions, where model predictions are the sum of components representing contributions coming from each input variable. \cite{Strumbelj2010shapley} proposed an approach for how to achieve this for a general model $\hat{f}$. The authors made the analogy between variables in a model and players in a cooperative game, where the joint prediction of variables in the model is seen as the payoff achieved by the players of the game, e.g.\ a football team winning a match.\\
The situation of the game already has a general solution which is given by the {\it Shapley value} attributed to each players (\citet{shapley1953shap}). This can be written as 
%
\begin{eqnarray}
\hat{f}(x_i)  \,&=&\, \phi_0\,+\,\sum_{k=1}^{n}\,\phi_{k}\big(x_i;\hat{f}\big)\,\equiv\,\Phi_1(x_i), \qquad \text{with} \label{eq:shap}\\
\phi_{k}\big(x_i;\hat{f}\big) \,&=&\, \sum_{x'\,\subseteq\,\mathcal{C}(x)\setminus k} \frac{|x'|!(n-|x'|-1)!}{n!}\,\big[\hat{f}(x_i|x'\cup \{k\}) - \hat{f}(x_i|x')\big]\,, \label{eq:shap2}
\end{eqnarray}
where $\mathcal{C}(x)\setminus\{k\}$ is the set of all possible variable combinations (coalitions) of $n-1$ variables when excluding $k$. The combinatorial weighting factor $|x'|!(n-|x'|-1)!/n!$ sums to one over $\mathcal{C}(x)$. Eq.\ \ref{eq:shap2} can be interpreted as the marginal contribution of variable $k$ to all possible coalitions excluding it. Model Shapley values have a set of appealing properties. In particular, they are the unique class of additive value attributions which is locally accurate (or efficient), respects missingness (the null player), is symmetric, has strong monotonicity, and, importantly, is linear (\citet{shapley1953shap,young1985shap,Strumbelj2010shapley}).\\
Equations\ \ref{eq:shap} \& \ref{eq:shap2} do not account for the case when variables jointly contribute to model predictions, i.e.\ when they are dependent or interact. This situation can be addressed by using the more general Shapley-Taylor decomposition proposed by \citet{mukund2019shapley_taylor}: the discrete set derivative of model $\hat{f}$ at point $x_i$ with respect to the set $x'$ conditioned on $x''$ is defined as
%
\begin{equation}\label{eq:set_der}
\delta_{x'} \hat{f}(x_i|x'') \,\equiv\, \sum_{x'''\,\subseteq\, x'} (-1)^{|x'''|-|x'|}\;\hat{f}(x_i|x'''\cup x'')\,.
\end{equation}
The case $|x'|=1$ corresponds to the bracket in (\ref{eq:shap2}). Let $h\leq n$ denote the maximal order of interaction terms we consider, then the Shapley-Taylor components for variables and their interactions up to order $h$ at $x_i$ is
%
\begin{equation}\label{eq:shap_taylor_ind}
\phi_h\big(\hat{f}, x_i\,|\,x'\big)\,=\,
\begin{cases}
      \delta_{x'} \hat{f}(x_i\,|\,\emptyset) & \quad\text{if}\quad |x'|<h\,,\\
      \frac{h}{n}\sum_{x''\subseteq \mathcal{C}(x) \setminus x'} \frac{\delta_{x'} \hat{f}(x_i\,|\,x'')}{\binom{n-1}{|x''|}} & \quad\text{if}\quad |x'|=h\,.
    \end{cases}
\nonumber
\end{equation}
Terms of order strictly smaller than $h$ are given by the set derivative (\ref{eq:set_der}) relative to the empty set, i.e.\ they are the net of interactions accounting for variable dependencies. We call those {\it bare components}. Terms of order one ($|x'|=1$) are the variable {\it main effects}. The full decomposition of model predictions up to order $h$ takes the form
%
\begin{equation}\label{eq:accuracy_2}
\hat{f}(x_i)\,=\,\phi_0\quad + \sum_{x' \subseteq \mathcal{C}(x),\,|x'|\leq h} \phi_h\big(\hat{f}, x_i\,|\,x'\big)\,=\, \sum_{k\in\{0,x'\}}\phi_{k;\,h}(\hat{f},x_i)\,\equiv\,\Phi_h(x_i)\,,
\end{equation}
with $\phi_0$ being the same intercept as in (\ref{eq:shap}). The second sum generalizes summation over variables to that over Shapley-Taylor components. We suppress the expansion order $h$ in what follows if not explicitly needed. Statistical models usually do not allow for missing inputs, i.e.\ $|x'|<n$. We address this by integrating out variables excluded from the model over a background $x_{\mathrm{bg}}$. The intercept $\phi_0$ is then the expected predicted value over $x_{\mathrm{bg}}$, i.e.\ $\phi_0=\mathbb{E}[\hat{f}(\emptyset)]$. This provides the reference value against which each Shapley component is measured. Thus, the choice of the background data is important for the interpretation of  Shapley components, which we will discuss below.\footnote{The formal properties of Shapley values, our computational approach, and a comparison of the Shapley-Taylor decomposition of model predictions and the the Taylor expansion of a analytical function are given in the Online Appendix.}

\section{Shapley value-based estimation and inference}
\label{sec:main}
\vspace*{-0.3cm}

We present our main theoretical results where the three subsections discuss estimation, inference, and signal and bias assessment, respectively. 

\subsection{Shapley estimation}
\label{sec:shapley_eff}

The coefficient concept from the linear model is, by definition, not applicable to models nonlinear in parameters. Instead we propose the use of the Shapley value components $\phi_{x'}(x_i;\hat{f})$ to estimate {\it local attributions} coming from the variable component $x'$ to model predictions of $\hat{f}$ at $x_i$. Over $R$ cross-fitting realizations $\phi_{x'}^s$, these can be written as
%
\begin{equation}\label{eq:effect_single}
\phi_{i,x'}^R\,\equiv\,\frac{1}{R}\sum_{s=1}^R \, \phi_{x'}^s\big(x_i;\hat{f}\big)\,.
\end{equation}
We can aggregate this over the full input space, or any subspace of interest, to give an {\it average attribution} stemming from $x'$,
%
\begin{equation}\label{eq:effect_agg}
\bar{\phi}_{x'}^R\,=\,\mathbb{E}_{\Omega}\big[\phi_{i,x'}^R\big] = \frac{1}{m}\sum_{i=1}^m \, \phi_{i,x'}^R\,.
\end{equation} 
\noindent While Equations\ \ref{eq:effect_single} \&\ \ref{eq:effect_agg} provide local and average attributions as measured by the generally nonlinear model $\hat{f}$, it is not clear if Shapley values are a desirable approach to do so. For instance, they are not the only way to decompose model predictions. Despite their appealing properties, we provide further results to motivate their use.
\begin{lemma} (analytical continuity I) \label{math:ana_conti_1}
The Shapley decomposition $\Phi$ of a model $\hat{f}$ linear in parameters $\theta$, $\hat{f}(x)=x\theta$, is the model itself with $\Phi=\Phi_1$ (Eq.\ \ref{eq:shap}), and $\phi_k^{\mathrm{lin}}=(x_k-\bar{x}_k)\hat{\theta}_k$ (Corollary\ 1 in \cite{lundberg2018shap}). The \hyperref[app:proof_ana_conti_1]{proof} is given in the Appendix.
\end{lemma}
\noindent Hence, the Shapley decomposition of the linear model is well known and variable attributions are directly proportional to the estimated coefficients $\hat{\theta}$ because of their constancy. Thus, model Shapley values can be seen as an extension of the coefficient concept when moving from the linear to a nonlinear model setting.\\
We next connect error consistency ({\it Assumption\ 1}), with estimation consistency of the true DGP by looking into the properties of Shapley values from universal approximators.
\begin{theorem}
\label{math:shap_consist}
(Shapley value consistency) The Shapley decomposition $\Phi$ of a model $\hat{f}$ converges component-wise and uniformly in probability to the Shapley decomposition $\Phi^{\star}$ of the true DGP. The \hyperref[app:proof_shap_consist]{proof} is given in the Appendix.
\end{theorem}

\noindent Theorem\ \ref{math:shap_consist} says that model Shapley values return the true contributions of variables to the DGP asymptotically, i.e.\ they are asymptotically unbiased. This means that they can be used to uncover the true but unknown DGP of interest. We will later provide ways to assess the convergence process and the trust we can have in Shapley estimates from different models.

\subsubsection{Application: Shapley estimation for heterogeneous treatment effects}
\label{sec:shap_treat}
%
Machine learning models have been extensively used to estimate treatment effects, which we take as our guiding example, and to show how Shapley value estimation can contribute to this literature. We consider the potential outcomes framework \citep{rubin1974estimating} where subjects either receive a treatment or not, i.e.\ $t_i\in\{0,1\}$ with $P(t_i=1)\in (0,1)$, leading to either $y_{i}^0$ or $y_{i}^1$, respectively. We assume unconfoundedness, i.e.\ that the treatment assignment $t_i$ is independent of the potential outcomes $y_{i}^t$ conditioned on the observables $z_i$, $\{y_{i}^0,y_{i}^1\}\perp t_i\,|\,z_i$. Setting $x_i=(t_i,z_i)$, the expected treatment effect can be written as $\tau(x_i) = \mathbb{E}[y_{i}^1 - y_{i}^0\,|\,z_i ]$.\\
The fundamental dilemma in the potential outcomes framework is that we only observe either $y_{i}^0$ or $y_{i}^1$ but never both. Moreover, treatment effects may be heterogeneous, e.g.\ there may be interactions of the treatment with the controls $z$ affecting $y$. We can use the Shapley-Taylor decomposition (\ref{eq:accuracy_2}) to investigate this. Without loss of generality, we set $h=2$ considering main and two-variable interaction effects. We then can derive the second-order {\it Shapley treatment function},\\
\begin{eqnarray}
\hat{\tau}(x_i) & = & \mathbb{E}\big[\hat{y}_i^1\hspace*{-.1cm}-\hspace*{-.1cm}\hat{y}_i^0|z_i\big] = \mathbb{E}\big[\hat{y}_i^1|z_i\big]\hspace*{-.05cm}-\hspace*{-.05cm}\mathbb{E}\big[\hat{y}_i^0|z_i\big] = \Phi_2(t\hspace*{-.1cm}=\hspace*{-.1cm}1,z_i)\hspace*{-.05cm}-\hspace*{-.05cm}\Phi_2(t\hspace*{-.1cm}=\hspace*{-.1cm}0,z_i)\label{eq:tf_row1}\\
          & = & \bigg[\sum_{k=0}^n\phi_{i,k}^{t=1} + \sum_{k,l;k>0,k>l}\phi_{i,kl}^{t=1}\bigg] - \bigg[\sum_{k=0}^n\phi_{i,k}^{t=0} + \sum_{k,l;k>0,k>l}\phi_{i,kl}^{t=0}\bigg]\label{eq:tf_row2}\\
          & = & \bigg[\phi_{i,t}^{t=1} + \sum_{k;k\notin\{0,t\}}\phi_{i,tk}^{t=1}\bigg] - \bigg[\phi_{i,t}^{t=0} + \sum_{k;k\notin\{0,t\}}\phi_{i,tk}^{t=0}\bigg]\label{eq:tf_row3}\\
          & = & \phi_{i,t}^{t=1} + \sum_{k;k\neq t}\phi_{i,tk}^{t=1} \;\equiv\; \phi_t + \sum_{k;k\neq t}\phi_{i,t*k}\,. \label{eq:tf_row4}
    \vspace*{-0.3cm}
\end{eqnarray} 
The second row (\ref{eq:tf_row2}) inserts the definition of the Shapley-Taylor decomposition where single and double indexed terms correspond to the main effects ($|x'|=1$) and pairwise interactions ($|x'|=2$), respectively. Any term not involving the treatment $t$ cancels out in the third row, including the intercept $\phi_0$.\\
Going to the forth row, we set the background data against which variable coalitions are evaluated to $x_{\mathrm{bg}}=(t=0,z)\equiv x_{\mathrm{bg}}^0$, i.e.\ the control group, or a representative subsample or summary of the untreated population. This has two consequences for for the treatment function. First, all terms in the second bracket of (\ref{eq:tf_row3}) vanish when going to (\ref{eq:tf_row4}) due to the missingness property of Shapley values. Second, the first terms $\phi_t$ in (\ref{eq:tf_row4}), which is constant across subjects, is an estimate of the average treatment effect (ATE).\\
Eq.\ \ref{eq:tf_row4} is appealing for several reasons. On the one hand, it decomposes the treatment effect into terms with intuitive interpretations: $\phi_{i,t*k}$ measures how the treatment varies alongside control $k$. This information may allow for the testing of hypotheses for treatment channels, which can be helpful to improve external validity of experimental interventions. On the other hand, the treatment function (\ref{eq:tf_row4}) can be directly estimated in a supervised learning setting predicting the response $y$ using off-the-shelf implementations of commonly used machine learning models. We label this the {\it direct} approach. Approaches based on the construction of orthogonalized scores are called the {\it indirect} approach. However, the treatment function (\ref{eq:tf_row4}) is easily derived for indirect approaches as well by applying the Shapley-Taylor decomposition to the predicted treatment effects.\\
Finally, we will consider the average treatment effect on the treated (ATT) for our empirical case study. This is given by
%
\begin{equation}\label{eq:att}
ATT\big(\hat{f},\Omega\big)\,=\,\mathbb{E}_{\Omega}\big[\hat{\tau}(x)\,|\,t=1\big]\,=\,\mathbb{E}_{\Omega}\Big[ \phi_t + \sum_{k;k\neq t}\phi_{i,t*k}\,\big|\, t=1 \Big]\,.
\end{equation}
We can obtain the ATE from (\ref{eq:att}) by swapping the treatment assignment label of the control group to estimate the potential, but unobserved, outcomes from the treatment function, and then taking the average over the full sample.

\subsection{Shapley estimation uncertainty}

We have so far provided individual or mean point estimates from nonlinear approximators. We next quantify the estimation uncertainty around those estimates. Here, we need to consider two sources of variation, conventional sampling, and sample split uncertainty due to cross-fitting.
%
\begin{algorithm}
    \caption{training bootstrap estimation}\label{algo:train_test}
    \begin{algorithmic}
    	\REQUIRE $\hat{f}$ (model type), $x$ (data), $B$ (number of bootstrap iterations), $K$ (number of cross-fitting folds)
    	\FOR{$s$ = $1$ to $K$}
        \STATE $\cdot$ split $x$ into $x_{\mathrm{train}}^s$ and $x_{\mathrm{test}}^s$ using cross-fitting
        \FOR{$b$ = $1$ to $B$}
                \STATE $\cdot$ draw bootstrap sample $x_b^s$ from $x_{\mathrm{train}}^s$
                \STATE $\cdot$ initialise a model $\hat{f}$
                \STATE $\cdot$ train $\hat{f}$ using $x_b^s$
                \STATE $\cdot$ perform Shapley decomposition $\Phi^b(x_{\mathrm{test}}^s;\hat{f}_b)$
        \ENDFOR
      \ENDFOR
      \STATE $\cdot$ form bootstrap estimates, e.g.\ $\phi_{i,x'}^B=\frac{1}{B}\sum_{b=1}^B \, \phi^b_{i,x'}$ for variable coalition\\component $x'$
      \STATE $\cdot$ calculate confidence intervals from bootstrap set $\{\phi^{b}_{i,x'}\}$ by method of choice
      \STATE $\cdot$ test hypothesis of interest, e.g.\ $\mathcal{H}^0_{x'}:\phi_{i,x'}^{\star}=0$
    \end{algorithmic}
\end{algorithm}
%
While nonasymptotic estimation bounds have been derived for specific models and applications \citep{Farrell2021deep}, we aim for a general computation approach. Sampling uncertainty of our estimates cannot be addressed at the stage of extracting the Shapley components, but needs to be done before model training. We therefore propose {\it training bootstrap estimation} for the derivation of confidence bounds.\footnote{This approach is similar to the one used in \cite{Cook2021boot} to quantify uncertainty around partial dependency plots, while no formal analysis has been presented there.} The estimation and inference procedure follows Algorithm\ \ref{algo:train_test}.
The validity of bootstrap estimates hinges on $\sqrt{m}$ convergence of $\hat{f}$, which is guaranteed by the training and test approach presented in Section\ \ref{sec:train_test_setting}. We can then derive the following results. 

\begin{theorem}\label{math:training_bs} {\it (training bootstrap consistency)}: Let $\hat{f}$ be a model  and $\hat{f}_b$ a realization trained on a bootstrap sample $x_b$. If $\hat{f}$ converges to $f$ (error consistency), so does $\hat{f}_b$. The \hyperref[app:proof_training_bs]{proof} is given in the Appendix.
\end{theorem}
\noindent Now, combining Theorem\ \ref{math:shap_consist} and Theorem\ \ref{math:training_bs} with the efficiency and linearity properties of Shapley values, we arrive at 
\begin{corollary} \label{math:shapley_bs} {\it (Shapley bootstrap consistency)}: Let $\hat{f}_b$ be a bootstrap realization of $\hat{f}$ trained on a bootstrap sample $x_b$ with Shapley decomposition $\Phi_b$. If $\hat{f}_b$ converges to $f=\Phi^{\star}$ (Theorem\ \ref{math:training_bs}), then $\Phi_b\rightarrow\Phi^{\star}$, and $var(\Phi_b)\rightarrow var(\Phi)$ of $\hat{f}$, both component-wise, as $m\rightarrow\infty$. The \hyperref[app:proof_shapley_bs]{proof} is given in the Appendix.
\end{corollary}
\noindent Corollary\ \ref{math:shapley_bs} states that bootstrap inference is asymptotically valid for our Shapley estimates. Its finite-sample precision will depend on the concrete setting. While sample size is an important factor here, it is hard to give specific guidance. However, we can reduce some of this uncertainty by 

\begin{proposition}
\label{math:prop_Shap_clt}
(Shapley central limit theorem) Let $\phi_{i,x'}(\hat{f})$ be observation-level Shapley estimates for a variable coalition $x'$, then the sampling distribution of the mean,  $\bar{\phi}_{x'}$, converges in distribution to a Gaussian. In particular, let $\bar{\phi}_{x'}^{\star}$ be the true mean with sample standard deviation $\sigma_{x'}^{\star}$, then 
%
\begin{equation}\label{eq:prop_Shap_clt}
\sqrt{m}\,\bar{\phi}_{x'}\;\; \underset{\mbox{\scriptsize $ m\rightarrow\infty$}}{\text{\Large\ensuremath \longrightarrow}}\;\; \mathcal{N}\big(\bar{\phi}_{x'}^{\star},\sigma_{x'}^{\star}\big)\,.\nonumber
\end{equation}
The \hyperref[app:proof_prop_Shap_clt]{proof} is given in the Appendix.
\end{proposition}

\noindent Proposition\ \ref{math:prop_Shap_clt} means that we obtain relatively tight bounds around mean estimates as the sample size increases. However, uncertainty remains about the estimation of the mean value for finite samples, since it does not need to be unbiased. This relates to potential biases in nonlinear estimation, which we will address in the next section.
Before this, we address sample split uncertainty, which may arise from cross-fitting or other techniques based on repeated estimation. To jointly account for sampling and sample split uncertainty in the cross-fitting setting, we would have to form $B \times R$ estimates (each involving $K$ folds). This can become computationally too demanding quickly. We approach the problem of joint estimation uncertainty using $B + R$ estimations by making the following approximation to confidence bounds. 

\begin{proposition}
\label{math:prop_ci}
(Sample split confidence intervals) Let $[\phi^B_{\mathrm{low}},\phi^B_{\mathrm{high}}]$ and $[\phi^R_{\mathrm{low}},\phi^R_{\mathrm{high}}]$ be confidence intervals at some level $\gamma$ from independent estimations, such as those resulting from bootstrap estimation (Algorithm\ 1) and sample split variation, respectively. Both intervals are median-centered at zero without loss of generality, then the joint confidence interval at level $\gamma$ is bounded as 
%
\begin{equation}\label{eq:prop_ci}
[\phi^{\mathrm{joint}}_{\mathrm{low}},\phi^{\mathrm{joint}}_{\mathrm{high}}]\;\subseteq\;[\phi^B_{\mathrm{low}}+\phi^R_{\mathrm{low}},\phi^B_{\mathrm{high}}+\phi^R_{\mathrm{high}}]\,.
\end{equation}
The \hyperref[app:proof_prop_ci]{proof} is given in the Appendix.
\end{proposition}

\noindent The right-hand side of Equation\ \ref{eq:prop_ci} provides a conservative bound assuming that large variations in the estimation of $\phi^B$ coincides with large variations of $\phi^R$. However, when equating these quantities with sampling and sample split estimates, respectively, we often observe that the latter has considerably smaller variation than the former, i.e.\ $\phi^R_{\mathrm{high}}-\phi^R_{\mathrm{low}} \ll \phi^B_{\mathrm{high}}-\phi^B_{\mathrm{low}}$, such that right-hand side of (\ref{eq:prop_ci}) provides a practical adjustment. We will see in the empirical case study that such adjustments accounting for split uncertainty are mostly small, especially for aggregate measures where variation from the observation level cancels out.

\subsection{Shapley regressions}
\label{sec:shapley_reg}

Estimation with linear models is often unbiased, e.g.\ under the conditions of the Gauss-Markov theorem. However, unbiased estimation is difficult to achieve in the general nonlinear setting. For instance, nonlinear problems quickly lead to non-convex optimization problems where estimation outcomes can vary with the model and optimization algorithm being used. As a consequence, different statistical learning models, even from the same model family, may learn different signals in finite samples despite their universal approximator properties. This means that Shapley value estimates may differ across models.\\
We address this by providing a simple parametric test with an intuitive asymptotic theory to assess the trust the modeler can have in Shapley component estimates from a particular model. By assumption, the Shapley decomposition of a model is aligned with the target, i.e.\ $\mathbb{E}[\Phi(x_i)]=y_i$. Based on this observation, we formulate the {\it Shapley regression} 
%
\begin{equation}
\label{eq:shap_reg}
y_i\,=\, \Phi(x_i)\hat{\beta}^S + \epsilon_i' \,=\, \sum_{k\in\{0,x'\}}\phi_{k}(\hat{f},x_i)\,\hat{\beta}^S_k + \epsilon_i'\,,
\end{equation}
with $\mathbb{E}[\epsilon_i'|\Phi(x_i)]=0$.\footnote{The term Shapley (value) regression has been used to address multi-collinearity in linear regression settings (see \cite{lipovetsky2001shap}). We do not see risk for confusion with the current unrelated setting.} Based on the key assumption of learning (\ref{eq:err_consist}) we can derive 

\begin{theorem}\label{math:sr_consist} {\it (Shapley regression asymptotics)}: Let $\hat{f}$ be a universal approximator and $\Phi$ its Shapley value decomposition (\ref{eq:accuracy_2}), then the true values of the components of $\beta^S$ for the Shapley regression (\ref{eq:shap_reg}) are either $\beta^S_k=0$ or $\beta^S_k=1$ for all $k\in\{0,x'\}$. The \hyperref[app:proof_sr_consist]{proof} is given in the Appendix.
\end{theorem}

\noindent The interpretation of Theorem\ \ref{math:sr_consist} is that a component, e.g. a single variable entering $\hat{f}$, is either part of the true DGP ($\beta^S_k=1$) or not ($\beta^S_k=0$).\footnote{Including the intercept in the regression is a notational convenience (as for linear regression models). It assures that we obtain the same asymptotic values for all $k\in\{0,x'\}$. Excluding $k=0$ leads to $\beta^S_0\rightarrow\phi_0$.} The latter case means that this component is pure noise in the problem at hand and not part of the DGP. These two cases can be statistically differentiated by testing $\hat{\beta}_k^S$ against the null hypothesis
%
\begin{equation}\label{eq:null}
\mathcal{H}^{0}_{k}(\Omega)\;:\;\{\beta_k^S\leq 0\,\big|\,\Omega\}\,,\,k\in\{0,x'\}\,.
\end{equation}
If (\ref{eq:null}) is rejected, we say that there is an alignment of component $\phi_{k}$ with the target, i.e.\ a signal stemming from this component. A difference to the conventional linear case is that hypothesis tests, such as against $\mathcal{H}^{0}$, will likely be more sensitive to the region $\Omega$ over which they are evaluated.  That is, only {\it local} statements about significance  can be made due to the potential nonlinearity of the model. If we reject $\mathcal{H}^{0}_{k}(\Omega)$, we accept the alternative hypothesis
%
\begin{equation}\label{eq:one}
\mathcal{H}^{1}_{k}(\Omega)\;:\;\{\beta_k^S=1\,\big|\,\Omega\}\,,\,k\in\{0,x'\}\,.\nonumber
\end{equation}
However, $\hat{\beta}_k^S$ may be far away from unity. We say that a component $\phi_{k}$ has been learned {\it robustly} if $\hat{\beta}_k^S\approx 1$ by some criterion. This can be statistical, like $\hat{\beta}_k^S=1$ being located centrally within the estimator distribution, or be taken to be a distance measure deemed close enough to one.\\
\begin{figure}[h!]
\centering
  \includegraphics[width=1.\linewidth]{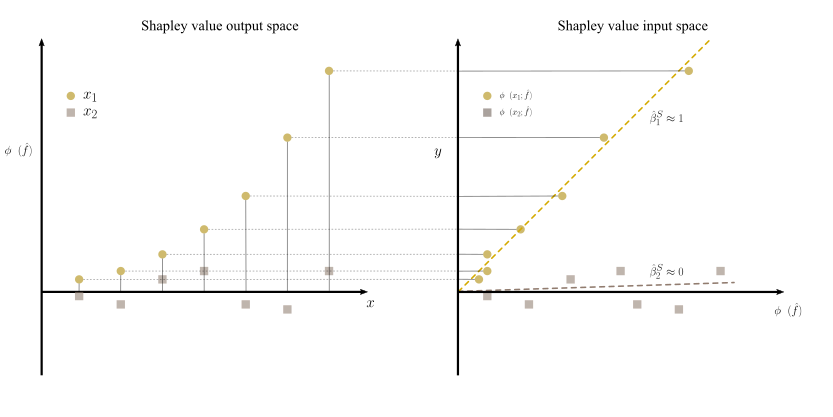}
  \vspace*{-1.5cm}
  \caption{\small{The principle behind Shapley regression (\ref{eq:shap_reg}): Shapley values project the learned functional forms of prediction components on the left-hand side (Shapley value output space) into a linear space with respect to the target space (right-hand side), where the true coefficient values $\beta^S$ can either be zero (noise) or one (signal).}}
  \label{fig:sr_principle}
\end{figure}
\noindent The concept behind Shapley regressions is illustrated in the stylized example shown in Figure\ \ref{fig:sr_principle}. We consider the problem $y=f(x_1)+\eta=\hat{f}(x_1,x_2)+\epsilon$. That is, the true DGP depends only on the variable $x_1$ (circles), while we model it using the pair $(x_1,x_2)$. For instance, we may have a (wrong) hypothesis in mind connecting $x_2$ (squares) to the DGP. Furthermore, we assume for the simplicity of presentation that the two variables could only enter the DGP additively, but not necessarily linearly, and that there is no intercept ($\phi_0=0$). This gives the full Shapley decomposition $\hat{f}(x_1,x_2)=\phi_{x_1}+\phi_{x_2}$.\\
After model fitting, the decomposition for some test predictions looks like the left-hand side of Figure\ \ref{fig:sr_principle}. We see that the learned functional form for variable $x_1$, $\phi_{x_1}$, shows an upward sloping nonlinear relationship, while $\phi_{x_2}$ does not exhibit patterns with values scattered around zero. The right-hand side of Figure\ \ref{fig:sr_principle} plots the target against the Shapley components which, by construction, absorb the nonlinearities of $\hat{f}$ (Shapley input space). The regression (\ref{eq:shap_reg}) now measures the alignment of each component with the target $y$ in this space, where we test $(\beta^S_1,\beta^S_2)$ against $\mathcal{H}^{0}$ from (\ref{eq:null}). We have $\hat{\beta}^S_k\approx 1$, i.e.\ $\hat{f}$ has learned the information from $x_1$ well, and we can say with high confidence that $x_1$ contributes to the underlying DGP. On the contrary, $\hat{\beta}^S_k\approx 0$, i.e.\ there is no clear alignment of the signal learned from $x_2$ with the target at this stage of learning. This means that either $x_2$ is pure noise, i.e.\ not actually part of the DGP, or its contribution to the DGP is badly measured and we may need more data to learn a signal coming from $x_2$.\\ 
The hypothesis (\ref{eq:null}) corresponds to the standard null in the linear regression setting, because Shapley values absorb the sign of model components such that only positive a $\hat{\beta}^S_k$ is indicative of a learned signal. This becomes clearer from 
\begin{lemma}
\label{math:prop_ana_conti_2}
(analytical continuity II) The Shapley regression (\ref{eq:shap_reg}) for a linear model $\hat{f}=x\theta$ is identical to the model itself, i.e.\ $\hat{\beta}^S=1$ with equivalent inference results. The \hyperref[app:proof_ana_conti_2]{proof} is given in the Appendix.
\end{lemma}
\noindent Lemma\ \ref{math:prop_ana_conti_2} says that a Shapley regression does not contribute anything new on top of the original model if this is a linear regression. As soon as we move away from the linear model, the $\hat{\beta}^S$ may differ from one due to incomplete learning in finite samples. This also means that tracking $\hat{\beta}^S$ for different samples sizes can be used to gauge the state of learning of different components $\phi_{x'}$. This brings us to

\begin{theorem}\label{math:ml_bias} {\it (unbiased estimation)}: Let $\phi_{x'},\; x' \in \mathcal{C}(x)$ be a bare component of a Shapley decomposition of a model $\hat{f}$ for points $x_i\in\omega\subseteq\Omega$, and $\phi_{x'}^{\star}(x_i)$ the corresponding true values from the DGP $f$.  Then, $\phi_{x'}$ is an unbiased estimate from $\hat{f}$ if $\hat{\beta}^S_{x'}=1$, such that $\mathbb{E}[\phi_{x'}]=\mathbb{E}[\phi_{i,x'}^{\star}],\,\forall x_i\in\omega$. Furthermore, the $\phi_{x'}$ is an unbiased estimate everywhere if $\hat{\beta}^S_{x'}=1, \forall \omega \subseteq \Omega$. The \hyperref[app:proof_ml_bias]{proof} is given in the Appendix.
\end{theorem}\vspace*{.2cm}
\noindent Theorem\ \ref{math:ml_bias} provides simple conditions for when we can have trust in the Shapley value estimates from a model, and how general this trust can be. The regression (\ref{eq:shap_reg}) estimates the $\hat{\beta}^S_k$ across the whole data set $x$. However, different distributions of $\phi_k$ may lead to the same $\hat{\beta}^S_k$. Having $\hat{\beta}^S_k\approx 1$ suggests that at least the central values of $\phi_k$ are well estimated and we may say that this component has been learned robustly. Now, if $\hat{\beta}^S_k\approx 1$ across any meaningful subregion of the whole input space, we have good alignment between Shapley components and the target everywhere, and we can say that these estimates are unbiased everywhere, i.e. corresponding to their asymptotic values.\\
It is important to consider bare components, i.e.\ those net of the interactions with other terms in the Shapley-Taylor decomposition, because this separates their estimation from the influences of other terms. Otherwise, the measurement of $\hat{\beta}^S_k$ would be affected by those interactions, for instance, because the level of $\phi_k$ could still change due to factors unrelated to the component of interest.\\
We can make the following qualitative statement. Conditioned on $\mathcal{H}^0_k$ being rejected, the relative magnitude of $\hat{\beta}^S_{k}$ for a bare component carries some information about the model's state of learning. We say that a model overestimates (underestimates) the effect from $k$ if $\hat{\beta}^S_k$ is smaller (bigger) than one. That is because the Shapley regression shrinks (inflates) the contributions of $\phi_{k}$ within each model prediction relative to the asymptotic limits ($\phi_k^{\star}, \beta^S_k=1$).\\
The Shapley regression (\ref{eq:shap_reg}) is based on generated regressors $\phi$ (\citet{Pagan1984generator}). Therefore, inference with regard to $\hat{\beta}^S$ is valid under two conditions.  First, the estimation of the coefficients $\hat{\beta}^S$ must be independent from the estimation of $\phi$. This is achieved by the i.i.d. assumption and standard sample splitting approach used in statistical learning. Models are optimized on the training set. Shapley value estimation and inference, and the estimation of $\hat{\beta}^S$, are done on an independent test set.\\
The second condition for valid inference is that nonparametric convergence of $\phi$ is at least of the rate $\sqrt{m}$. Both conditions are met by the sample splitting and test approach described in Section\ \ref{sec:train_test_setting} ({\it Assumption\ 2}). Because of this, we also can consider the uncertainty of the estimation of Shapley values in the previous section separate from that of the coefficients $\hat{\beta}^S$. The former can be treated as a variable transformation entering the estimation of the latter. As the final part of our theory, we link the components $\phi_k$ to the corresponding coefficients $\hat{\beta}^S_k$.
%
\begin{corollary}
\label{math:prop_test_hierarchy}
(test hierarchy) Let $\phi_{x'},\; x' \in \mathcal{C}(x)$ be a component of a Shapley decomposition of a model $\hat{f}$. If we reject $\mathcal{H}^0:\{\phi^{\star}_{x'}=0\}$ within some region $\omega\subseteq\Omega$, we can also reject $\mathcal{H}^0:\{\beta^S_{x'}\leq 0\}$ in that region. The \hyperref[app:proof_prop_test_hierarchy]{proof} is given in the Appendix.
\end{corollary}
\noindent Corollary\ \ref{math:prop_test_hierarchy} says that, if $\phi_k$ is bounded away from zero, it also can be said to contribute to the true DGP of interest. This is of practical relevance in situations where we measure a clear signal stemming from $\phi_k$ but cannot reject the null with respect to $\hat{\beta}^S_k$. This results comes, however, with two caveats. One, the regional dependence on $\omega\subseteq\Omega$ is important as the values of $\phi_k$ may well include the zero within $\omega$. Two, rejecting only $\mathcal{H}^0(\phi^{\star}_k)$ but not $\mathcal{H}^0(\beta^S_k)$, we cannot make a statement about potential estimation bias with regard to $\phi_k$.

\section{Applications}
\label{sec:applications}

We consider two case studies estimating heterogeneous treatment effects: a simulation and a real-world experiment. The details of implementation for both are given in the Online Appendix alongside additional results. 

\subsection{A numerical experiment with unknown treatment interaction}
\label{sec:sim}

\begin{figure}[!h]
\centering
  \includegraphics[width=.725\linewidth]{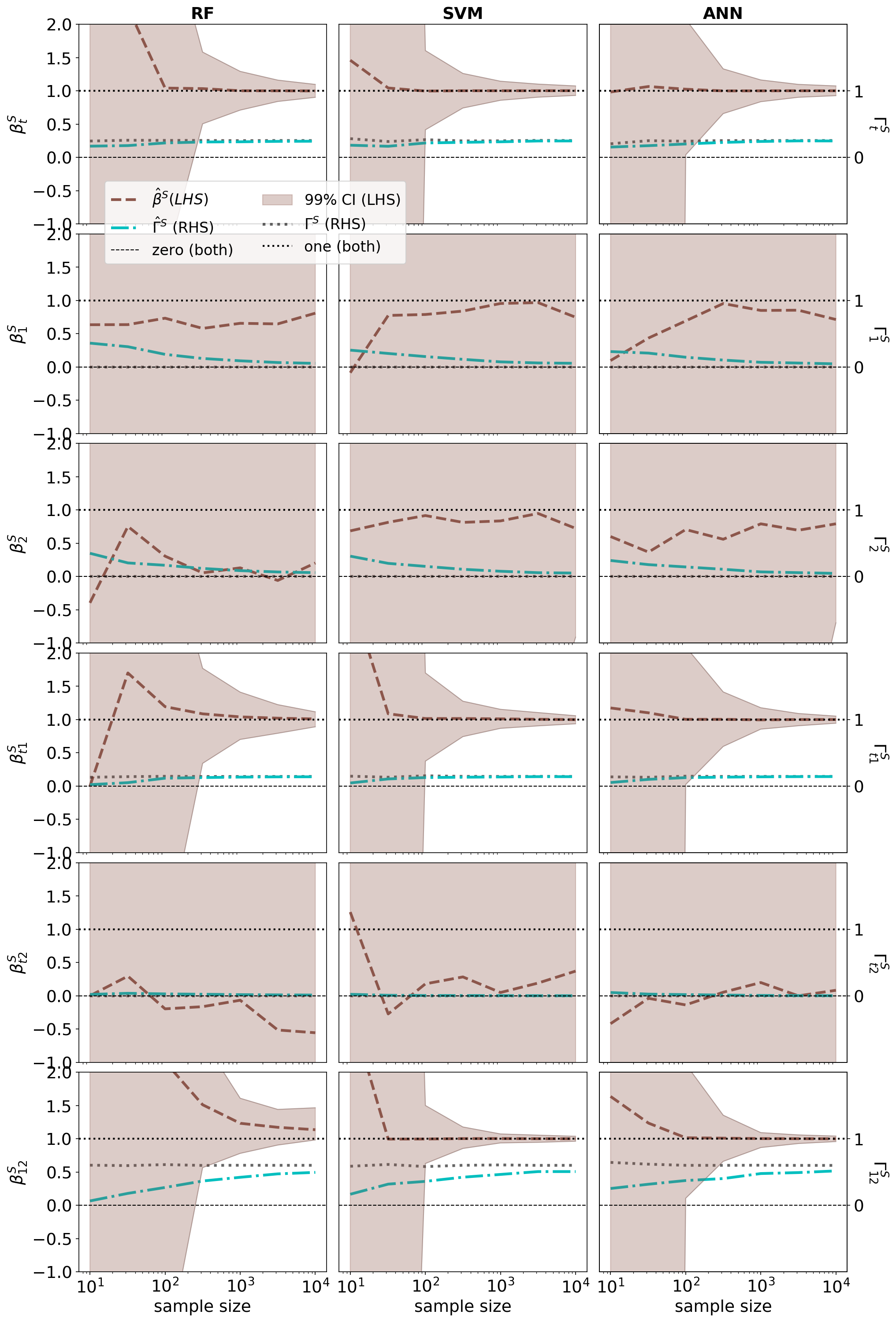}
  \caption{\small{Inference analysis on simulated DGP (\ref{eq:sim_dgp}) using RF, SVM and ANN (columns) for learning selected real and spurious components (rows). Left-hand side axes: Shapley regression coefficients $\hat{\beta}^S$ (dashed lines) and 99$\%$\ confidence intervals (shaded areas). Right-hand side axes: True ($\Gamma^S$, dotted lines) and learned ($\hat{\Gamma}^S$ dashed-dotted lines) Shapley predictive shares. We excluded the terms $\phi_{0}$ and $\phi_{t12}$ for better presentation, for which the results are analogous to the other spurious components.}}
  \label{fig:inf_sim}
\end{figure}

Let $x=(t,z_1,z_2)$ with $t\sim\mathcal{B}(0.5)\in\{0,1\}$ a treatment drawn from a fair Bernoulli distribution, and $z_k\sim\mathcal{N}(0,1),\,k\in\{1,2\}$, covariates sampled form a standard normal distribution. We consider the DGP 
%
%
\begin{equation}\label{eq:sim_dgp}
y\;=\;f_t(x;\alpha)+\eta \;=\; \alpha_1 t + \alpha_2 t z_1 + \alpha_3 z_1 z_2 + \alpha_4 +\eta\,,
\end{equation}
with $\eta\sim\mathcal{N}\big(0,0.1\,\sigma^2(f_t)\big)$ an irreducible noise component drawn from a normal distribution centered at zero and a standard deviation of $10\%$ of $f_t$. The coefficients are set to $\alpha=(1, 1, 1, 0)$. The DGP has a heterogeneous treatment component ($\alpha_2$), while the ATE is set by $\alpha_1$ as $\mathbb{E}[z_1]=0$. Without {\it a priori} knowledge of $f_t$, the heterogeneous treatment component is challenging to estimate as $z_1$ does not only interact with the treatment but also with the covariate $z_2$. We are interested in recovering the DGP (\ref{eq:sim_dgp}) from noisy observations $(x,y)$. We will also investigate the the asymptotic properties of estimation by considering different sample sizes, and look at the heterogeneous treatment effect in more detail. We simulate multiple realizations of $f_t$ for different sample sizes and use off-the-shelf implementations of different statistical learning models (RF, SVM, ANN) for cross-validation, and testing. Using the latter, we decompose all model predictions into the Shapley-Taylor decomposition, which to full order ($h=3$) takes the form
%
\begin{eqnarray}
\hat{f}_t(x_i) \,&=&\, \phi_{0} +\phi_{i,1}+\phi_{i,2}+\phi_{i,1*2}\nonumber\\
&+& \phi_{t}+ \phi_{i,t*1}+\phi_{i,t*2}+\phi_{i,t*1*2}  \label{eq:tf_psi} \\
\,&=&\, \phi_{0}+\phi_{i,1}+\phi_{i,2}+\phi_{i,1*2} \;+\;  \hat{\tau}(x_i)\,. \label{eq:tf_psi_2}
\end{eqnarray} 
The components $\phi_{0}$ (intercept), $\phi_{1}$, $\phi_{2}$, $\phi_{t2}$, and $\phi_{t12}$ are spurious as they are not present in the DGP (\ref{eq:sim_dgp}). Models should not learn those components, while they may me measured to be non-zero due to imperfect learning and the presence of noise. The second row (\ref{eq:tf_psi_2}) singled out the treatment function $\hat{\tau}(x_i)$ from (\ref{eq:tf_row4}). If the focus is on the investigation of treatment effects, and not, say, on recovering to full DGP, the consideration of only the components of $\hat{\tau}(x_i)$ suffices.\\
We will focus on the full DGP here, and inference is performed in two steps. First, we perform a Shapley regression (\ref{eq:shap_reg}) on (\ref{eq:tf_psi}) to identify components which are likely part of the true DGP. We then look for treatment heterogeneity based on the component $\phi_{t1}$ after it has been identified to contribute to the DGP. The results for the first step are summarized in Figure\ \ref{fig:inf_sim} with the sample size on the horizontal axis in all panels.
Each row corresponds to a summand in (\ref{eq:tf_psi}) and the columns correspond to the models (RF, SVM, ANN). The left-hand side vertical axes refer to the coefficients $\hat{\beta}^S$ (dashed lines). The shaded areas are the $99\%$-confidence intervals. The right-hand side vertical axis shows the estimated predictive share of that Shapley component $\hat{\Gamma}^S$ (dashed-dotted lines), among all components, relative to the corresponding truly realized share (dotted lines).\\
Only three of the six shown components of (\ref{eq:tf_psi}) converge to the signal value $\beta^S=1$, namely those corresponding to $\phi_{t}$, $\phi_{t1}$ and $\phi_{12}$. These are precisely the components actually present in $f_t$. The sample dependence of the spurious components does not show patterns of convergence. The confidence intervals of the latter components are always overlapping with zero, such that we cannot discern them from noise. Furthermore, their predictive shares $\Gamma^S$ convergence to zero with increasing sample size. All components' learned shares converge to their true values. These findings are in line with Theorem\ \ref{math:shap_consist}, and we can confidently distinguish between true and spurious components at sample sizes above a few hundred to a thousand depending on the model.
\begin{figure}[h!]
\centering
  \includegraphics[width=.85\linewidth]{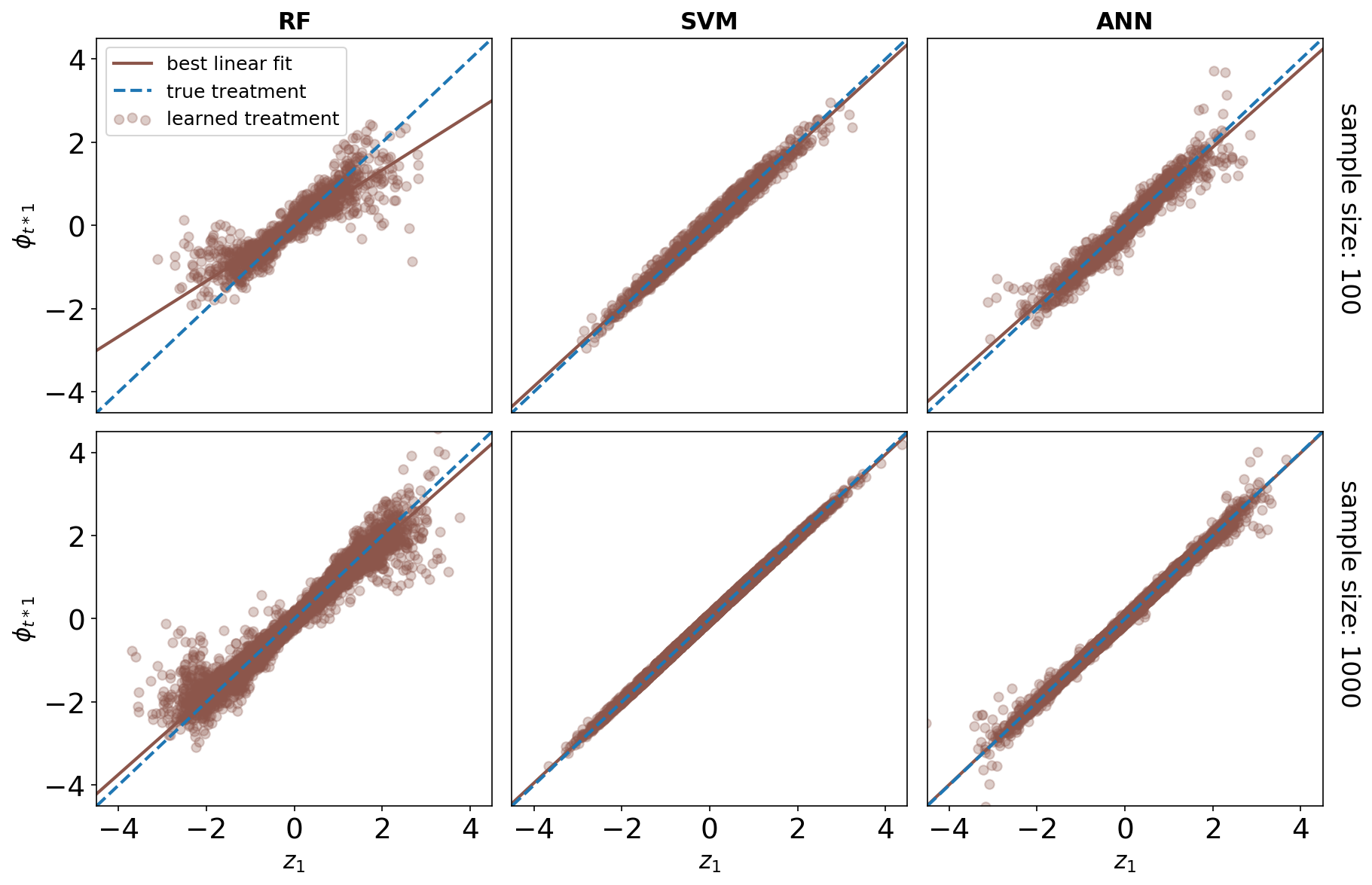}
  \caption{\small{True (dashed lines) versus learned (dots) treatment interaction effects $\phi_{t*1}$ for RF, SVM and ANN (columns) for different sample sizes: 100 (upper row), 1000 (lower row). Best-fit linear treatment functions are given by the solid lines.}}
  \label{fig:hetero_sim}
\end{figure}
We next investigate the learning of treatment heterogeneity by considering the term $\phi_{t1}$ in more detail. Latest from a sample size of 1000, this component is highly significantly and robustly estimated to be part of the true DGP by all models. We compare the learned Shapley values for that component with the observed inputs. This is shown in Figure\ \ref{fig:hetero_sim} which depicts the extracted Shapley component on the vertical axes versus the input values $z_1$ for the treated ($t=1$) for all models (columns) and two sample sizes (rows). The solid line in each panel shows the best linear fit to the estimated treatment relations, which can be compared to the dashed lines representing the true treatment.\\
There are clear differences between models and between samples sizes. The RF seems to have difficulty of learning this term. This is not surprising, because tree-based models are not well suited for modeling smooth functions which are not aligned with the variable axes. On the contrary, the SVM learns an almost perfect representation of the heterogeneous treatment term. Even for a sample size of only 100, the estimated linear treatment function (solid line) almost perfectly coincides with the true treatment (dashed line). This means that we have successfully uncovered this part of the true DGP with the tools presented in this paper.

\subsection{A real-world experiment}
\label{sec:bank_rct}

We revisit parts of the analysis in \cite{BHOLAT2018,BHOLAT2019}. The authors ran a randomized control trial to investigate the effects of different information treatments on the comprehension of a monetary policy statement from the Bank of England by the general public. We focus on the `relatable summary' (treatment) compared to the more technical plain text summary (control) of the statement. This treatment explained economic conditions and monetary policy decisions via a combination of graphical and simple textual descriptions, which related to common experiences like grocery shopping. The understanding of either monetary policy communication was assessed by the same comprehension test. The relatable summary lifted scores substantially. Out of a maximal score of seven on a [0-7] discrete scale, the average score after reading the plain text summary was $2.53$, while the score for the relatable summary was $3.80$. That is, the treatment lifted scores by about $1.27$ points, or by about $50\%$ relative to the control.\\
Direct estimation is based on the specification, $y_i\,=\,\mathit{score}_i \,=\, \hat{f}\big(t_i,z_i;\theta\big) + \epsilon_i$. The treatment $t$ is one for subjects who saw the relatable summary, and zero for the plain text summary, with $m=1066$ ($m_{t=0}=538$, $m_{t=1}=628$). The vector $z_i$ contains demographic controls, especially age which we will focus in more detail below. We use cross-fitting and nested cross-validation to train and test the same off-the-shelf models (RF, SVM, ANN) as in simulation study. We also estimate a set of models following the indirect approach: causal forests (CRF; \cite{athey2016forest,wager2018treatment}), and the X-learner version of RF, SVM, and ANN from \cite{kunzel2019meta}, where each of the prediction steps within the learner is performed by the respective base learner.
We construct the treatment function (\ref{eq:tf_row4}), for which we expand model predictions to third order ($\Phi_h(x_i),\,h=3,\,x_i=(t_i,z_i)$). This guarantees that the main ($h=1$) and pairwise interactions ($h=2$) effects are the net of higher-order terms. The treatment function can be written in an abbreviated form as 
%
\begin{equation}\label{eq:tf_abbrev}
\hat{\tau}_i=\hat{\tau}(x_i)=\phi_{t}+\phi_{i,t*\mathit{age}}+\phi_{i,t*\mathit{income}}+\phi_{i,t*\mathit{gender}}+\phi_{i,t*\mathit{resid}}\,,
\end{equation}
where terms of the form $\phi_{t*z_k,i}$ are the estimated treatment contributions from the pairwise interaction of the treatment with the demographic characteristic $z_{k}$ of individual $i$. The term $\phi_{t*\mathit{resid},i}$ is the sum of the pairwise interactions of the treatment with other controls and all other higher-order terms. We can represent the treatment function as any combination of sums of components due to the linearity of Shapley values.
%
\begin{table}[ht]
\resizebox{\textwidth}{!}{%
\begin{tabular}{c|c|ccc|c|ccc||c}
term & quantity & \multicolumn{3}{c|}{direct} & \multicolumn{4}{c||}{indirect} & direct \\
\hline
  & model & {\bf RF}  & {\bf SVM} & {\bf ANN} & {\bf CRF} & {\bf X-RF} & {\bf X-SVM} & {\bf X-ANN} & {\bf OLS}\\
\hline
 $\hat{\tau}$ & $ATT$ & 1.28 & 1.25 & 1.21 & 1.19 & 1.31 & 1.40 & 1.32 & 1.33\\
 &  $CI_{95}^{\mathrm{adj}}$ & \small{[1.17,1.38]} & \small{[1.12,1.39]} & \small{[0.90,1.43]} & \small{[1.07,1.31]} & \small{[1.21,1.39]} & \small{[1.29,1.51]} & \small{[1.20,1.43]} & \small{[1.24,1.42]}\\
 &  $CI_{95}^{\phantom{adj}}$ & \small{[1.18,1.37]} & \small{[1.16,1.35]} & \small{[1.04,1.35]} & \small{[1.09,1.29]} & \small{[1.22,1.39]} & \small{[1.30,1.50]} & \small{[1.22,1.41]} & \small{[1.24,1.42]}\\
\hline
 $\phi_{t}$ & $ATT$ & 1.28 & 1.25 & 1.21 & 1.19 & 1.32 & 1.43 & 1.31 & -\\
  &  $CI_{95}^{\mathrm{adj}}$ & \small{[1.16,1.39]} & \small{[1.11,1.38]} & \small{[0.91,1.43]} & \small{[1.07,1.30]} & \small{[1.22,1.43]} & \small{[1.31,1.55]} & \small{[1.17,1.46]} &
\end{tabular}
}
\caption{\small{ATT over training bootstrap realizations for the different models and (sample-split adjusted) confidence intervals at the $95\%$ level for $\hat{\tau}$ (upper part) and the bare treatment component $\phi_{t}$ (lower part). The results for the linear model (OLS) are obtain from the corresponding estimates of the treatment regression coefficient with the same inputs and using the same sample splits after averaging the results from the $K$ folds.}}
\label{tab:att}
\end{table}

\subsubsection{Treatment effect estimation}

We first investigate aggregate and heterogeneous treatment effects for the different models, before moving to an example of treatment interaction channels in the next subsection. The ATT obtained from all models is shown in Table\ \ref{tab:att}, with the linear estimate given for reference in the last column. Looking at the upper $\hat{\tau}$-part, we see that most statistical learning models estimate the ATT well. The ANN and CRF, however, learn a somewhat subdued treatment effect. Again with the exception of these two models, the confidence intervals of all the machine learning estimates $CI_{95}^{\mathrm{adj}}$ strongly overlap among each other and with that of the linear model, and they have comparable widths.\\
To asses practical aspects of our theory, we make two further comparisons at this point. First, we look at the bare treatment term $\phi_{t}$ in the treatment function (\ref{eq:tf_abbrev}). This should give the same ATT estimates if the treatment and control groups have the same demographic sample statistics, and if the latter (or a summary of it) has been used as the Shapley value background. This is indeed the case when comparing the upper and lower parts of Table\ \ref{tab:att}. Both the estimated ATT and their confidence intervals are almost identical in almost all cases.
Second, we investigate the importance of the sample split adjustment of confidence intervals for the ATT estimates. We previously noted that split variation will mostly cancel out for aggregated effects. This is the case when comparing the adjusted $CI_{95}^{\mathrm{adj}}$ and the unadjusted $CI_{95}$ intervals in the $\hat{\tau}$-part of Table\ \ref{tab:att}. This potentially allows the modeler to save on computational resources if only aggregate estimates are of interest.

\begin{figure}[t!]
\centering
  \includegraphics[width=.8\linewidth]{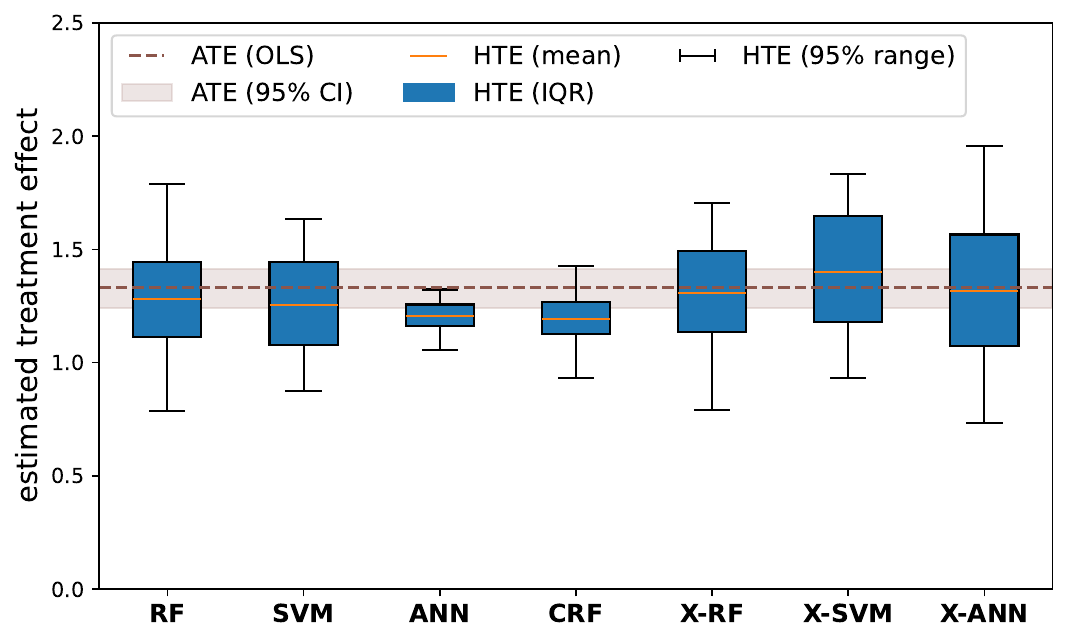}
  \caption{\small{Distributions of $\hat{\tau}_i$ for different models using box plots: mean (center line), interquartile range (IQR; box), $95\%$ quantile range (whiskers). The linear ATE estimate is shown for reference (horizontal dashed line) with the corresponding $95\%$ confidence intervals (shaded area) from Table\ \ref{tab:att}.}}
  \label{fig:hte_box}
\end{figure}
\noindent We next look at the distributions of estimated treatment effects from the different models. These are shown in Figure\ \ref{fig:hte_box} taking the mean of the bootstrap realizations. Each box plot depicts the distribution of treatment effects for a model. We see that there is substantial variation between models. The ANN and CRF show relatively little variation around their central estimates (center line), while the estimated range of the X-ANN is comparatively wide. Generally, the treatment effect distributions learned by the X-learners are centered around the ATE estimate of the linear model, which provides some trust in those models.\\
We investigate the quality of the signals learned by the different models by testing the alignment of the learned distributions of treatment effects with the outcome using a Shapley regression, where we only consider the treatment effect and an intercept,
%
\begin{equation}\label{eq:shap_reg_simple}
y_i\,=\,c\,+\,\hat{\tau}_i\,\hat{\beta}^S_t\,+\,\epsilon_i'\,.
\end{equation}
It is instructive to consider two different samples here. The full sample combining the treated and the control group, and the sub-sample of the treated only. The Shapley regressions for the two samples investigate different questions. The asymptotic limit of the coefficient $\hat{\beta}^S_t$ is one if the treatment has some effect, which we established  with the results in Table\ \ref{tab:att}. This means that, when considering the treated only, a coefficient of one means perfect learning across the treated sample. We, thus, would have high confidence in the learned distribution of treatment effects. On the contrary, using the full sample, a $\hat{\beta}^S_t=1$ merely means that the model distinguishes well between the treated and the control group. That is, it has learned the ATT well, while there still can be degeneracy about the distribution of the treatment effect around the correct central value (as in Figure\ \ref{fig:hte_box}). As such, perfect learning ($\hat{\beta}^S_t=1$) for the treated subsample implies perfect learning for the full sample, but not vice versa. We note that distribution degeneracy within the treated sample with $\hat{\beta}^S_t=1$ is still possible. This can be tested by evaluating $\mathcal{H}^1_t$ over all subsets of the input space $\Omega$, achieved via adequate sampling when implemented. There is no degeneracy if $\hat{\beta}^S_t=1$ everywhere and we have recovered the true distribution (Theorem\ \ref{math:ml_bias}).
\begin{center}
\begin{table}[ht]
\resizebox{\textwidth}{!}{%
\begin{tabular}{c|c|ccc|c|ccc}
sample & statistic &{\bf RF}  & {\bf SVM} & {\bf ANN} & {\bf CRF} & {\bf X-RF} & {\bf X-SVM} & {\bf X-ANN} \\   
\hline
full & $\hat{\beta}_{t}^S$ & $0.89$  & $1.02$  & $0.91$  & $1.01$ & $0.95$ & $0.97$ & $0.84$ \\
 & $CI_{95}^{\mathrm{adj}}$ & \small{[0.79,1.00]} & \small{[0.92,1.13]} & \small{[0.73,1.03]} & \small{[0.90,1.12]} & \small{[0.87,1.04]} & \small{[0.89,1.06]} & \small{[0.73,0.96]} \\
 & $p(\mathcal{H}^0)$ & 0.00 & 0.00 & 0.00 & 0.00 & 0.00 & 0.00 & 0.00 \\
\hline
treated & $\hat{\beta}_{t}^S$ & $0.33$  & $0.99$  & $0.30$  & $0.24$  & $0.70$   &  $1.52$ & $0.39$\\
 & $CI_{95}^{\mathrm{adj}}$ & \small{[-0.33,0.94]} & \small{[-0.16,2.09]} & \small{[-0.76,1.50]}& \small{[-1.38,1.76]} & \small{[0.08,1.29]} & \small{[0.93,2.11]} & \small{[-0.12,0.89]}\\
 & $p(\mathcal{H}^0)$ & 0.15 & 0.04 & 0.31 & 0.37 & 0.02 & 0.00 & 0.06 \\
\end{tabular}
}
\caption{\small{Shapley regression coefficients of the comprehension score on the full treatment effect only including the treatment term and an intercept for the full sample (upper part) and the treated subsample (lower part). All statistics are obtained from the joint distribution of training bootstrap and cross-fitting estimates based on the inner $95\%$ percentiles of treatment Shapley values (outlier removal).}}
\label{tab:shap_main}
\end{table}
\end{center}
%
%
The results of this exercise are shown in Table\ \ref{tab:shap_main} which documents $\hat{\beta}_{t}^S$ for the different models and samples. Looking at the full sample (upper part), we see that all models learned a clear treatment signal. The estimated values are highly significant, and almost all are close to unity (robust learning), especially for the SVM and X-RF.\\
Looking at the treated subsample in the lower part of Table\ \ref{tab:shap_main}, we see that only three out of the seven models considered (SVM, X-RF, X-SVM) may be said to have learned a reliable distribution of treatment effects, in the sense that these are aligned with the observed outcomes. This observation is in line with Figure\ \ref{fig:hte_box}, where these models have comparable treatment effect distributions. The none-results for the ANN and the CRF also are in line with the rather compressed distributions in Figure\ \ref{fig:hte_box}. This indicates that these models could not well differentiate treatment heterogeneity.

\subsubsection{Treatment channels: the role of age}

\begin{figure}[h]
\centering
    \includegraphics[width=0.4\textwidth]{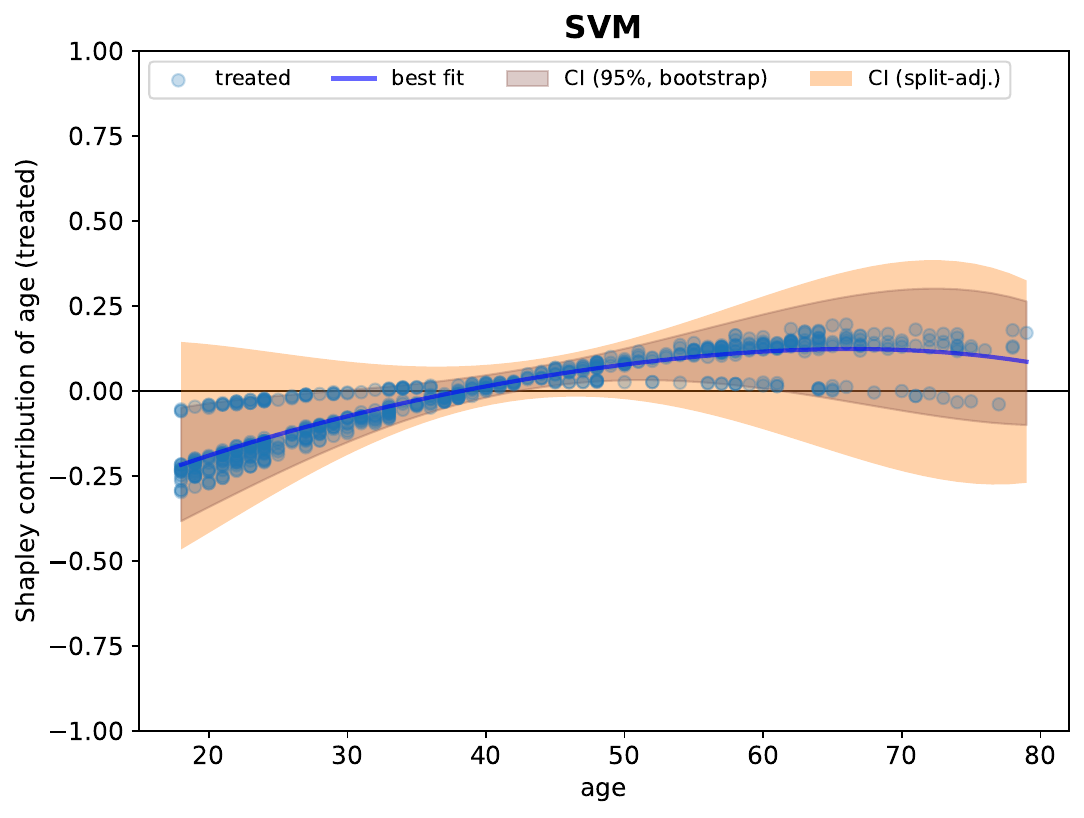} 
    \includegraphics[width=0.4\textwidth]{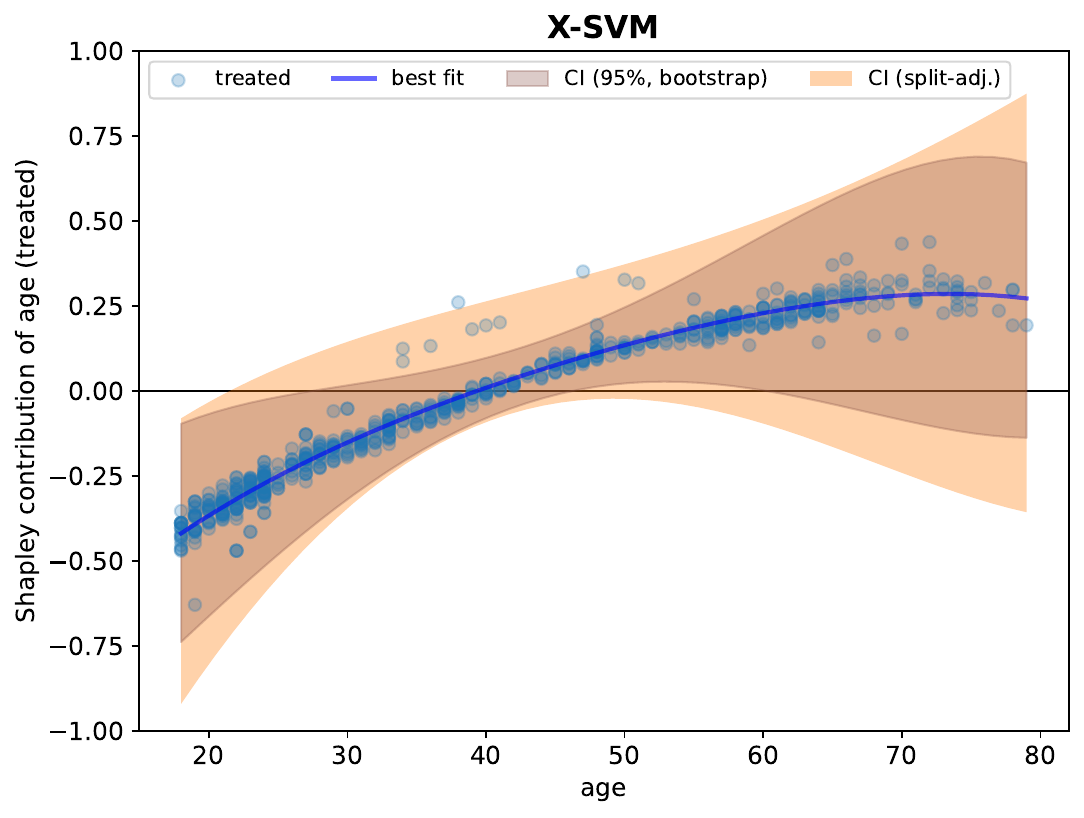}\\
    \includegraphics[width=0.4\textwidth]{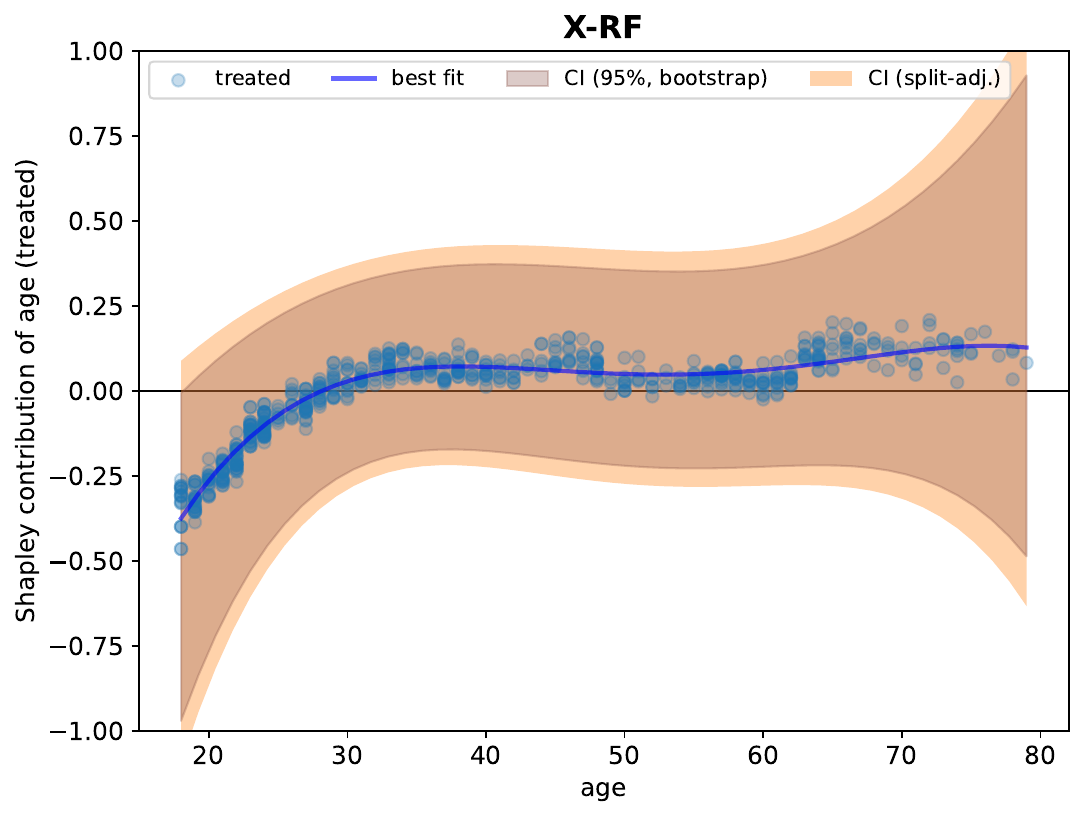} 
    \includegraphics[width=0.4\textwidth]{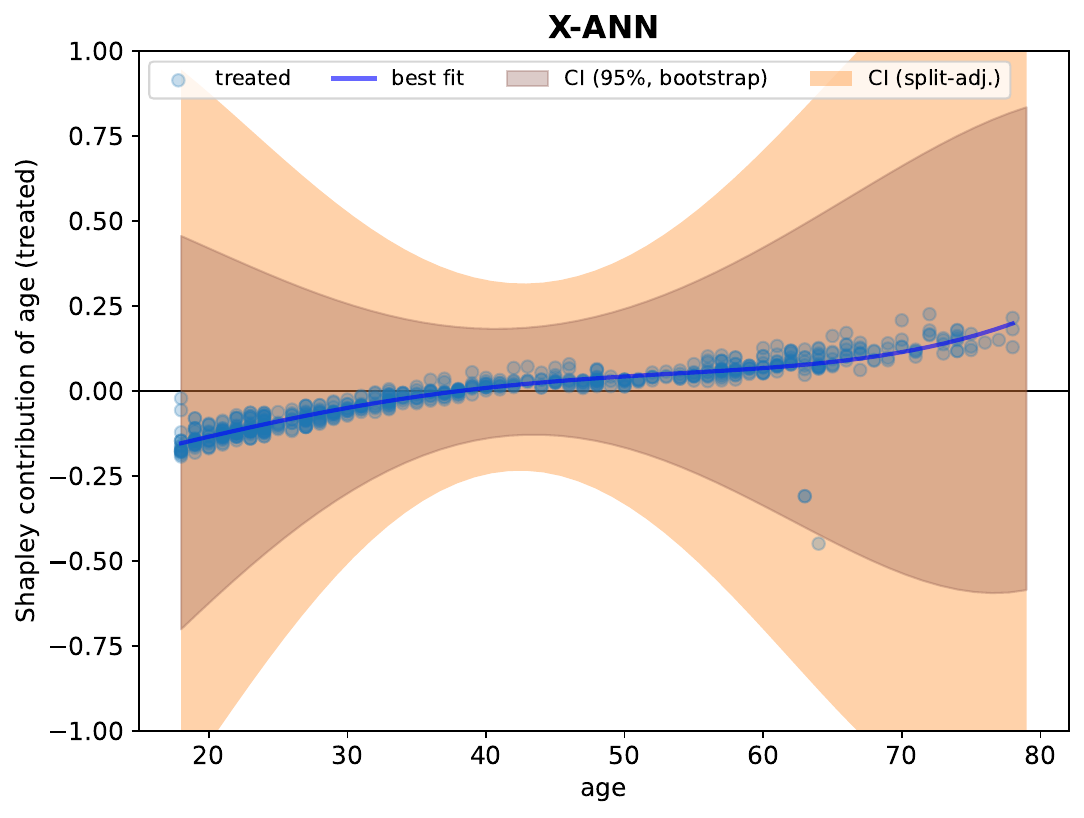}
  \caption{\small{Estimated treatment function component $\phi_{t*age}$ for different models as a function of age for the treated based on the training bootstrap distribution: individual means (dots), best fits (solid lines, degree-4 polynomials), $95\%$ confidence intervals (CI; inner), sample split adjusted CI (outer). Three individuals older than 80 have been excluded for clearer presentation.}}
  \label{fig:direct_treat_age}
\end{figure}

We investigate the role of age, which we measure to be a sizable sources of treatment heterogeneity. To do so, we analyze the Shapley interaction term $\phi_{t*age}$ in the treatment function (\ref{eq:tf_abbrev}). This measures the individual, potentially nonlinear, treatment contribution attributed to age relative to an individual of average age in the control group.\\
%
Figure\ \ref{fig:direct_treat_age} plots the learned treatment contributions from age for different models. In each panel, the horizontal axis is age and the vertical axis is $\phi_{t*age}$ for each individual in the treatment group (dots). The inner and outer bands respectively show the $95\%$ percentile bootstrap and sample-split adjusted confidence intervals. The best-fit lines approximate the learned functional forms.\footnote{The treatment-age functions for all models are given in the Online Appendix.}\\
We make several observations. First, all models learn a positive relationship between age and the effectiveness of treatment. Slopes of linear regressions applied to the data from each panel are all positive and highly significant. Second, the sample split adjustment of confidence bands can be important for an individual's estimate, despite these effects largely canceling out for aggregate estimates. The widths of the outer bands in Figure\ \ref{fig:direct_treat_age} relative to the inner bands are non-negligible, especially for the (X-)SVM. Third, despite the overall positive relationship between age and estimated treatment effectiveness, the confidence intervals for individuals of all ages and for most models overlap with the zero line. However, this does not mean that we cannot reject the null of no treatment effect by age for most individuals and models. To understand this better, and treatment functions more generally, we need to remember what the zero lines in Figure\ \ref{fig:direct_treat_age} mean. These are, by construction, the zero effects we expect to observe for individuals of the mean age in the control group, which again is the background against which we computed Shapley components. Empirically, this is about 41 years of age. Reassuringly, this also is about the point where the (X-)SVM models intersect the zero line; again providing confidence in these models. Thus, the zero-overlapping confidence intervals mean that, for most individuals and models, we cannot reject the null of no treatment effect with respect to those of average age. However, this will change with the reference value chosen for the Shapley value computation. We investigate differences in estimated treatment effects for different age groups by comparing the estimated treatment-age effects in the lower and upper quintiles of the age distribution. A one-sided $t$-test strongly rejects the null of equal means, again for all models.\\
However, what we do observe is large variation in the distribution of estimated treatment-age interactions together with different shapes of the best-fit treatment lines, especially by based model types like the RF and (X-)SVM. We can again quantify these differences using Shapley regressions of the form (\ref{eq:shap_reg_simple}), where we replace $\hat{\tau}_i$ with $\phi_{t*\mathit{age},i}$. The results for this exercise are shown in Table\ \ref{tab:shap_int} for the treated subsample.\\
Many models struggle to learn a relation between the treatment-age term and the outcome, since the null cannot clearly be rejected. In line with previous findings, the (X-)SVM models learn a highly significant relation. Taken together with their relatively smooth and narrow treatment functions in Figure\ \ref{fig:direct_treat_age} with the expected zero-intercepts, the evidence suggests that their qualitative features, e.g.\ the curves' comparable shapes, may be trusted. However, quantitative features, like estimated treatment-age values for individuals need to be treated with caution as $\hat{\beta}_{t*\mathit{age}}^S\gg 1$. These results also suggest that collecting more data, to the extent feasible, may be a fruitful way to achieve a better heterogeneous treatment effect estimation in this case.
\begin{center}
\begin{table}[ht]
\resizebox{\textwidth}{!}{%
\begin{tabular}{c|c|ccc|c|ccc}
sample & statistic & {\bf RF}  & {\bf SVM} & {\bf ANN} & {\bf CRF} & {\bf X-RF} & {\bf X-SVM} & {\bf X-ANN} \\   
\hline
treated & $\hat{\beta}_{t*age}^S$ & $0.92$  & $5.01$  & $3.00$  & $0.21$  & $1.65$   &  $2.79$ & $2.00$\\
 & $CI_{95}^{\mathrm{adj}}$ & \small{[-0.55,2.25]} & \small{[3.57,6.68]} & \small{[-2.52,8.81]}& \small{[-5.73,6.22]} & \small{[-0.11,3.13]} & \small{[2.07,3.72]} & \small{[-0.77,4.92]}\\
 & $p(\mathcal{H}^0)$ & 0.12 & 0.00 & 0.15 & 0.47 & 0.03 & 0.00 & 0.10 
\end{tabular}
}
\caption{\small{Shapley regression coefficients of the comprehension score on the treatment-age interaction effect only including this term and an intercept for the treated subset. All statistics are obtained from the joint distribution of training bootstrap and cross-fitting estimates based on the inner $95\%$ percentiles of treatment Shapley values (outlier removal).}}
\label{tab:shap_int}
\end{table}
\end{center}
%
Finally, we interpret the shapes of the learned (X-)SVM treatment curves in Figure\ \ref{fig:direct_treat_age}. They suggests that the information treatment is the more effective the older a person is, and that this effect levels off at between 60 to 70 years of age. Given that the treatment was meant to relate to lived experience, this suggests that this experience may help to understand economic concepts when presented in the right way, but also that there is a limit to this effect. These observations are in line with \cite{binder2024comms}, according to which age mostly has a positive relation with monetary policy knowledge. However, the authors also report negative relations in a minority of studies. The approaches presented here may allow to map such relationships with respect to demographic characteristics in more detail than was previously possible, which could reconcile these different results.

\section{Discussion}
\label{sec:conclusion}

We propose a generic estimation and inference framework for universal approximators which encompasses many models from the statistical learning literature. The approach consists of two steps after model training. First, the decomposition of model predictions into Shapley components of interest. These are our estimators for the generally nonlinear model. Second, inference on these components. This can involve either the testing of standard hypotheses of interest, e.g.\ if a component is likely to be different from zero, or the assessment of the quality of learning using Shapley regressions.\\
The latter can be done using linear Shapley regressions which are accompanied by a simple and intuitive theory. The true values of the resulting coefficients are either one or zero. These outcomes correspond to the two mutually exclusive cases that the respective and potentially nonlinear contributions measured by that Shapley components are part of the DGP in question, or that they are noise, i.e.\ not part of the DGP. For example, we may determine that a quantity we assumed to contribute to a process does actually not do so at a certain level of confidence. More generally, this allows us to uncover the unknown true DGP as we have shown in our numerical case study.\\
Furthermore, estimation and inference reduces to the well-known case of analyzing regression coefficients if the model is linear in parameters. Hence, the proposed framework can be seen as a `natural extension' of statistical inference when moving from the linear parametric to the nonlinear domain.\\
A crucial difference to the linear case is that estimation and inference is local, i.e.\ results are only valid within the region of the input space which has been investigated. For instance, a variable may be found to contribute to a DGP within some part of the input space, but not in others.\\
It is known in statistical learning that one often cannot know which model out of several will perform best, e.g.\ has the highest prediction accuracy, to address a particular problem \citep{Fernandez2014}. The mirror image of this for statistical inference is that we cannot know which model out of several nonparametric approximators will be best for estimation in a particular setting. This is precisely what we observed in our case studies, i.e.\ that we encountered large differences for some estimation outcomes between models, but that we may not have a prior to why this is the case. The framework presented here offers a versatile toolbox to quantify those differences. For instance, it allows us to assess the usefulness of different models for a particular estimation task based on estimation uncertainty or potential biases. In this way, we contribute to reconciling Breiman's ``two cultures'' \citep{breiman2001cultures}, the one of statistical rigor and the other of computational solutions, by allowing statistical inference without having to assume a stochastic model of the observed data.\\
Despite the estimation of treatment effects in the experimental setting being our guiding example, our framework is independent of a particular identification approach. This means that there is plenty of scope for future work to investigate nonparametric estimation and inference in different identification settings, but also settings with omitted variables, or with temporal dependencies in the DGP.

\section*{Appendix: proofs}
\label{sec:app}

\subsubsection*{Proof of Lemma\ \ref{math:superconv}}
\label{app:math:superconv}
\vspace*{-0.3cm}

If $K>\underline{K}$, the model $\hat{f}$ converges with a rate $\xi^{eff}_{ml} (\xi_{ml}, K)>\frac{1}{2}$ on the $\frac{m}{K}$ test partition. We can write the variance of a linear estimator based on $\hat{f}$ relative to an estimator converging with the parametric rate $\xi_p=\frac{1}{2}$, as

\vspace*{-0.8cm}
\begin{equation}
\frac{\text{Var}(E_{ml})}{\text{Var}(E_{p})} \sim \frac{m^{-2\xi^{eff}_{ml}}}{m^{-2\xi_p}} = m^{1-2\xi^{eff}_{ml}} = \frac{1}{m^{\delta}}\rightarrow 0,\quad\text{as }\,m \rightarrow\infty\;\text{with }\;\delta=2\xi^{eff}_{ml}-1>0\,. \pushQED{\qed}\qedhere\popQED
\end{equation}
\vspace*{-0.8cm}

\subsubsection*{Proof of Proposition\ \ref{math:ana_conti_1}}
\label{app:proof_ana_conti_1}
\vspace*{-0.3cm}

Observing that the intercept in a multiple linear regression is $x_0=\bar{y}-\bar{x}\hat{\theta}$ and that $\phi_0 = \hat{f}(\bar{x};\hat{\theta}) =\bar{y}$ for the linear model, we have 
\vspace*{-0.3cm}
\begin{equation}
\Phi(x) = \sum_{k=0}^n \phi_k^{lin} =  \bar{y} + \sum_{k=1}^n (x_k-\bar{x}_k)\hat{\theta}_k = x_0 + \sum_{k=1}^n x_k\hat{\theta}_k = x\hat{\theta}= \hat{f}(x;\hat{\theta})\,,
\end{equation}
with $\theta_0=1$. The properties of Shapley values for the $\phi_k$ can be easily verified. $\pushQED{\qed}\qedhere\popQED$
\vspace*{-0.3cm}
\subsubsection*{Proof of Theorem\ \ref{math:shap_consist}}
\label{app:proof_shap_consist}
\vspace*{-0.3cm}

Without loss of generality, let us assume that the columns of $x$ are independent, such that the Shapley-Taylor expansion (\ref{eq:accuracy_2}) reduces to the simple Shapley decomposition (\ref{eq:shap}). That is, there are only variable main effects and no interaction terms. For any $\delta>0$, we have
\vspace*{-0.5cm}
\begin{eqnarray}
0&=&\lim_{m \to \infty} \mathbb{P}\Big(\mathbb{E}\big[||f-\hat{f}\big|x||\big]>\delta\Big)=\lim_{m \to \infty} \mathbb{P}\Big(\mathbb{E}\big[||\sum_{k=0}^n\phi_k^{\star}(f)-\sum_{k=0}^n\phi_k(\hat{f})\big|x||\big]>\delta\Big)\nonumber\\
 &=&\lim_{m \to \infty} \sum_{k=0}^n\mathbb{P}\Big(\mathbb{E}\big[||\phi_k^{\star}-\phi_k\big|x||\big]>\delta\Big)\quad\Rightarrow\quad\phi_k\rightarrow\phi_k^{\star}\quad\text{as }\,m\rightarrow\infty\,.
\vspace*{-0.3cm}
\end{eqnarray}
The first equality is the key assumption of learning, the second uses the exactness property of Shapley values, while the last equality makes use of the independence of the components, which we assumed but do not need as an always-finite set of interaction terms can be added to the above expression. $\pushQED{\qed}\qedhere\popQED$

\vspace*{-0.3cm}
\subsubsection*{Proof of Theorem\ \ref{math:training_bs}}
\label{app:proof_training_bs}
\vspace*{-0.3cm}

Let $\hat{f}_m$ be a model trained on $m$ observations and $\hat{f}^b_m$ be a training bootstrap realization of the model. Then, the following holds for the central bootstrap estimate of $\hat{f}_m$. 
\begin{equation}
\mathbb{E}^B\big[\hat{f}_m\big]=\frac{1}{B}\sum_{b=1}^B\mathbb{E}\big[\hat{f}_m^b\big]=\mathbb{E}\big[\hat{f}_m\big]\rightarrow f,\quad\text{as } m\rightarrow\infty\,.
\end{equation}
Here we used that the expectation of the bootstrap trained model is the same as the one trained on the original sample and the key assumption of learning. $\pushQED{\qed}\qedhere\popQED$ 

\vspace*{-0.3cm}
\subsubsection*{Proof of Corollary\ \ref{math:shapley_bs}}
\label{app:proof_shapley_bs}
\vspace*{-0.3cm}

Let $\hat{f}^b_m$ be a training bootstrap realization of the model $\hat{f}_m$ trained on $m$ observations. Then, the following holds
\vspace*{-0.3cm}
\begin{equation}
\mathbb{E}\big[\Phi^b_m\big]=\mathbb{E}\big[\hat{f}^b_m\big]=\mathbb{E}\big[\hat{f}_m\big]\rightarrow f=\Phi^{\star},\quad\text{as } m\rightarrow\infty\,.
\end{equation}
Here we again used that the expectation of the bootstrap trained model is the same as the one trained on the original sample and the key assumption of learning. We also obtain convergence for the variance of bootstrap estimates $\Phi^b$ since $\hat{f}$ and as such $\Phi^b$ are square integrable. $\pushQED{\qed}\qedhere\popQED$ 

\vspace*{-0.3cm}
\subsubsection*{Proof of Proposition\ \ref{math:prop_Shap_clt}}
\label{app:proof_prop_Shap_clt}
\vspace*{-0.3cm}

Convergence to the mean value ($\bar{\phi}^{\star}_{x'}$) is given by Theorem\ \ref{math:shap_consist}. Because of the train-test setting (Assumption\ 2) and the assumption of finite variance of Shapley estimates, the conditions of the classical central limit theorem are fulfilled. That is, components $\phi_{i,x'}$ can be interpreted as realizations of independent random variables with finite variance, whose sampling mean converges to a normal distribution. $\pushQED{\qed}\qedhere\popQED$

\vspace*{-0.3cm}
\subsubsection*{Proof of Proposition\ \ref{math:prop_ci}}
\label{app:proof_prop_ci}
\vspace*{-0.3cm}

Without loss of generality, let the sampling distributions of $\phi^B$ and $\phi^R$ have both $q$ elements. We order both in ascending order and form the new random variable $\phi^{\mathrm{joint}}=\phi^B+\phi^R$ by the sum of ordered pairs. Then, at a given level $\gamma$, the confidence interval of $\phi^{\mathrm{joint}}$ is $CI_{max}(\phi^{\mathrm{joint}},\gamma)=[\phi^B_{\mathrm{low}}+\phi^R_{\mathrm{low}},\phi^B_{\mathrm{high}}+\phi^R_{\mathrm{high}}]$, i.e.\ the sum of bounds of the confidence intervals of the component variables. Since any other different relative ordering of $\phi^B$ and $\phi^R$ before forming $\phi^{\mathrm{joint}}$ can only reduce its variance, $CI_{\mathrm{max}}$ is the widest possible interval of $\phi^{\mathrm{joint}}$. $\pushQED{\qed}\qedhere\popQED$

\vspace*{-0.3cm}
\subsubsection*{Proof of Theorem\ \ref{math:sr_consist}}
\label{app:proof_sr_consist}
\vspace*{-0.3cm}

Without loss of generality, let the true DGP $f$ depend on a single variable $x_1$, and let us include two independent variables $x_1$ and $x_2$ in the universal approximator $\hat{f}$. That is, any signal coming from $x_2$ will be spurious (as in Figure\ \ref{fig:sr_principle}). Now, the Shapley decomposition of $\hat{f}$ takes the form $\hat{f}=\sum_{k=0}^3\phi_k$ with $\phi_0$ being the intercept. Then, 
\vspace*{-0.3cm}
\begin{equation}
\mathbb{E}[\hat{f}]=\mathbb{E}\big[\sum_{k=0}^2\phi_k\hat{\beta}^S_k\big]\quad\rightarrow\quad\phi^{\star}_0+\phi^{\star}_1=f,\quad\text{as }\;m\rightarrow\infty. 
\end{equation}

We used again Theorem\ \ref{math:shap_consist} for the limit on the right-hand side. A component-wise comparison implies that $\beta_k^S=1$ for $k \in \{0,1\}$, and $\beta^S_2=0$. Note that asymptotically $\mathbb{E}[\phi^{\star}_2]=0$ as well. However, one will almost surely measure a finite value for $\phi_2$ due to noise and imperfect learning while its mean converges to zero, such that the probability of wrongly rejecting $\mathcal{H}_2^0$ will tend to zero with increasing sample size. Hence, $\hat{\beta}_2^S\rightarrow 0$ and $\beta_2^S=0$. $\pushQED{\qed}\qedhere\popQED$

\vspace*{-0.3cm}
\subsubsection*{Proof of Lemma\ \ref{math:prop_ana_conti_2}}
\label{app:proof_prop_ana_conti_2}
\vspace*{-0.3cm}

The Shapley regression for the linear model is
\vspace*{-0.3cm}
\begin{equation}
\Phi(x)\hat{\beta}^S  = \hat{f}(x;\hat{\theta})\hat{\beta}^S  = x\, diag(\hat{\theta})\hat{\beta}^S = x\hat{\theta}.
\end{equation}
This follows from Proposition\ \ref{math:ana_conti_1} and the uniqueness of the coefficients $\hat{\theta}$ as solution to the convex least-squared problem. This can be made explicit for the OLS estimator. By setting $x\rightarrow x\, diag(\hat{\theta})\equiv x D_{\hat{\theta}}$, one obtains
\vspace*{-0.3cm}
\begin{equation}
\hat{\beta}^S = \frac{x D_{\hat{\theta}} y}{(x D_{\hat{\theta}})^T(x D_{\hat{\theta}})} = \frac{D_{\hat{\theta}}}{D_{\hat{\theta}}^2} \frac{X y}{x^T x} = D_{\hat{\theta}}^{-1}\hat{\theta} = 1_{n+1}\,.
\end{equation}
We can see that the above expression leads to the same inference results, as $se(\hat{\beta}^S)=se(\hat{\theta})/\hat{\theta}$ for the standard errors of the Shapley regression coefficients. $\pushQED{\qed}\qedhere\popQED$.

\vspace*{-0.3cm}
\subsubsection*{Proof of Theorem\ \ref{math:ml_bias}}
\label{app:proof_ml_bias}
\vspace*{-0.3cm}

By Theorem\ \ref{math:shap_consist}, $\mathbb{E}[\phi_{x'}]\rightarrow\mathbb{E}[\phi_{x'}^{\star}]$, as $m\rightarrow\infty$ with $x'\in\mathcal{C}(x)$. Assuming that no components in $\mathcal{C}(x)$ are spurious to the DGP $f$, we have $\beta^S_{x'}=1,\, \forall x'\in\mathcal{C}(x)$ and $x_i \in \omega\subseteq\Omega$. However, $\hat{\beta}^S_{x'}=1$ also is the asymptotic limit where $\mathbb{E}[\phi_{x'}\hat{\beta}^S_{x'}]=\mathbb{E}[\phi_{x'}^{\star}]$, and hence, $\mathbb{E}[\phi_{x'}]=\mathbb{E}[\phi_{x'}^{\star}]$, which corresponds to unbiased estimation. $\pushQED{\qed}\qedhere\popQED$

\vspace*{-0.3cm}
\subsubsection*{Proof of Corollary\ \ref{math:prop_test_hierarchy}}
\label{app:proof_test_hierarchy}
\vspace*{-0.3cm}

Without loss of generality, only a single variable $x_1$ enters the true DGP $f$, such that the model $\hat{f}$ only contains a single non-spurious Shapley component $\phi_1$, and $\hat{f}=\phi_0+\phi_1$ (with $\phi_0$ being the intercept). Furthermore, we set the background $x_{bg}$ such that $\phi_0=0$, and we assume rejections of the null hypothesis $\mathcal{H}^0(\omega):\{\phi^{\star}_1=0\}$, with $\omega\subseteq\Omega$, at some appropriate confidence level. We then have the following asymptotics for the corresponding Shapley regression (suppressing observation indices), 
\begin{equation}
\mathbb{E}[y]=\hat{\beta}_1^S\phi_1 \rightarrow \phi_1^{\star},\quad m\rightarrow\infty\,.
\end{equation} 
Because we rejected $\phi_1=0$, $\beta^S_1=1$ for the limit to old. $\pushQED{\qed}\qedhere\popQED$


\setstretch{1.5}

\bibliographystyle{aea}
\bibliography{Shapley_inference_v8_AE_arXiv}

\end{document}